\documentclass{article}

\PassOptionsToPackage{square,numbers,compress,sort}{natbib}

\usepackage[preprint]{neurips_2025}

\usepackage[utf8]{inputenc} 
\usepackage[T1]{fontenc}    
\usepackage{hyperref}       
\usepackage{url}            
\usepackage{booktabs}       
\usepackage{amsfonts}       
\usepackage{nicefrac}       
\usepackage{microtype}      
\usepackage{xcolor}         

\usepackage{amsmath,amsfonts,bm}

\def\eqref#1{equation~\ref{#1}}

\def\1{\bm{1}}

\def\rvh{{\mathbf{h}}}
\def\rvu{{\mathbf{i}}}

\def\rvp{{\mathbf{p}}}
\def\rvq{{\mathbf{q}}}

\def\rvu{{\mathbf{u}}}

\def\rvx{{\mathbf{x}}}

\def\rvz{{\mathbf{z}}}

\def\rmA{{\mathbf{A}}}
\def\rmB{{\mathbf{B}}}
\def\rmC{{\mathbf{C}}}
\def\rmD{{\mathbf{D}}}

\def\rmG{{\mathbf{G}}}
\def\rmH{{\mathbf{H}}}

\def\rmK{{\mathbf{K}}}

\def\rmQ{{\mathbf{Q}}}

\def\rmS{{\mathbf{S}}}

\def\rmV{{\mathbf{V}}}
\def\rmW{{\mathbf{W}}}

\def\rmZ{{\mathbf{Z}}}

\DeclareMathAlphabet{\mathsfit}{\encodingdefault}{\sfdefault}{m}{sl}
\SetMathAlphabet{\mathsfit}{bold}{\encodingdefault}{\sfdefault}{bx}{n}

\def\gD{{\mathcal{D}}}

\def\gL{{\mathcal{L}}}

\def\gO{{\mathcal{O}}}

\def\gX{{\mathcal{X}}}
\def\gY{{\mathcal{Y}}}

\def\sR{{\mathbb{R}}}

\newcommand{\E}{\mathbb{E}}

\usepackage{graphicx}
\usepackage{multirow}
\usepackage{colortbl}
\usepackage{wrapfig}
\usepackage{fontawesome5}
\usepackage{subcaption}
\usepackage{dsfont}
\usepackage{bbm}

\newcommand{\highblue}[1]{{\textbf{\color[RGB]{30, 85, 170}#1}}}
\newcommand{\highred}[1]{{\textbf{\color[RGB]{220, 20, 60}#1}}}
\def\eg{\textit{e.g.}}
\def\ie{\textit{i.e.}}
\newcommand{\rev}[1]{#1}
\newcommand{\dgv}[1]{#1}

\usepackage{algorithm}
\usepackage{algorithmic}

\title{Learning Mamba as a Continual Learner: Meta-Learning Selective State Space Models \\ for Continual Learning}

\author{Chongyang Zhao,~ Dong Gong\thanks{D. Gong is the corresponding author.}\\
University of New South Wales (UNSW Sydney)\\
{\texttt{\{chongyang.zhao, dong.gong\}@unsw.edu.au}}
}

\begin{document}

\maketitle

\begin{abstract}
Continual learning (CL) learns from a non-stationary data stream without storing or re-training on all seen samples. Meta-continual learning (MCL) casts CL as sequence prediction and meta-learns the continual learner itself as a sequence model, with Transformers as natural choices. However, despite decent performance, a Transformer learner relies on a linearly growing key-value cache to store all past representations, conflicting with CL’s objective of not storing all seen samples. Sequence models with a \emph{constant-size state}, e.g., linear-attention models and state-space models (SSMs), match CL's requirement by definition; however, with past samples compressed and mixed in the bounded state, such learners are harder to meta-learn, and earlier kernel-based ones performed poorly on MCL. We thus aim to obtain effective continual learners with constant-size states and make them work for MCL. By formulating the selective SSM for MCL, we propose \textbf{MambaCL}, with Mamba as the main model in practice. We meta-train the learner with a proposed selectivity regularization, which supervises the implicit associations within the compressed state and applies across constant-state models. Furthermore, we conduct a systematic empirical study of how Mamba and other constant-state models behave across various MCL scenarios, covering effectiveness, generalization, and model designs. The results highlight promising performance and strong generalization, demonstrating the potential of constant-state learners for efficient continual learning and adaptation.
\end{abstract}

\section{Introduction}
\dgv{Continual learning (CL) aims to efficiently learn and accumulate knowledge from a non-stationary data stream, without storing or re-training on all seen data \citep{de2021continual,wang2024comprehensive}. Given a stream $\gD_T=( (\rvx_1, y_1), ..., (\rvx_T, y_T) )$ of paired observations (\eg, images) and targets (\eg, class labels), CL learns one model $P_{\phi_t}(y|\rvx)$ to perform prediction for all tasks covered by the seen data $\gD_t$. For example, in class incremental learning (CIL)~\citep{rebuffi2017icarl,zhou2023deep}, new classes arrive incrementally and the model must recognize all seen classes. CL methods thus minimize the storage of historical data and constrain memory growth, ideally to a sub-linear rate \citep{de2021continual,ostapenko2021continual,wang2024self}.} 
\dgv{The key challenge is maintaining performance on past tasks while continually updating the model parameters $\phi_t$~\citep{de2021continual,wang2024comprehensive}: the learner must retain useful knowledge without retaining the data itself.} 
\par
\dgv{A continual learner turns a data stream into a sequence of models: at each step, $\phi_{t+1}=\texttt{optim-step}(\phi_t, \rvx_t, y_t)$, so each model is determined by the seen data $\gD_t$.
From this optimization viewpoint, learning and prediction together form a sequence prediction (SP) problem \citep{lee2023recasting,bornschein2024transformers}: predicting $y^\text{test}$ conditioned on the seen sequence and the query, $(\gD_t^\text{train}, \rvx^\text{test})$.
Rather than hand-crafting the update rule, meta-continual learning (MCL) \citep{lee2023recasting,son2024meta,bornschein2024transformers} meta-learns the continual learner itself: a sequence model $f_{\theta}()$ is trained across CL episodes to predict $y_{t+1}=f_\theta( (\gD_t, \rvx_{t+1}) )$, with the stream as the prediction context.
This parallels in-context learning (ICL) in large language models, which adapt to a task from examples in the context without updating weights \citep{brown2020language,garg2022can,min2021metaicl}; we pursue this ability for general tasks beyond text, \eg, episodes of image classification and regression.}

\par
Considering the strong sequence modeling ability, Transformers \citep{vaswani2017attention,touvron2023llama} are a natural choice for SP-based MCL \citep{lee2023recasting,son2024meta}. 
\dgv{A meta-learned Transformer stores a key-value pair for every seen sample in a \emph{key-value cache} and predicts by attending over \emph{all} of them; this retrieval-based learner performs well on CL and MCL benchmarks \citep{lee2023recasting,bornschein2024transformers}.
However, its state---the key-value cache \citep{katharopoulos2020transformers,lee2023recasting}---grows linearly with the stream, and each prediction scans the entire cache, leading to computation quadratic in the episode length.
Such a design is not ideal for a continual learner, which should not maintain and repeatedly access all seen samples: the cache amounts to storing the whole data pool rather than learning from it.
This raises a natural question: \emph{can we obtain a meta-learned continual learner that performs well and also has the state property CL favors, \ie, not retaining all seen samples?}}
 
\dgv{Sequence models with a \emph{constant-size state}---including linear-attention models, \eg, Linear Transformer (Linear TF) \citep{katharopoulos2020transformers} and Performer \citep{choromanski2020rethinking}, and state-space models (SSMs) \citep{gu2021efficiently,gu2021combining,gu2023mamba}---compress the history into a fixed-capacity state with linear computation complexity.
By definition, this format is what CL asks of a learner: knowledge is carried in a bounded state, and no per-sample storage is required. The constant state, however, also brings the difficulty: past samples are compressed and mixed within the fixed capacity, so the learner must continually select what to keep, and learning such a learner is correspondingly harder.
Earlier kernel-based linear-attention models were indeed found to perform poorly on MCL \citep{lee2023recasting}, and whether a constant-state learner can be made to work well has remained open.}

\par
\dgv{We thus aim to obtain effective continual learners with constant-size states---and to make them work for MCL.
We formulate the selective SSM for MCL and meta-learn the model as an online continual learner, an instantiation we call \textbf{MambaCL}, with Mamba \citep{gu2023mamba,dao2024transformers} as the main model in practice: its input-dependent selective state updates give the learner an explicit mechanism for choosing what to write into its bounded state.
Meta-training such learners is challenging, since the supervision must reach the associations between queries and the relevant past samples that are only implicitly retained in the compressed state.
We introduce a selectivity regularization to supply this guidance, relying on the connections among the selective SSM, linear attention, and softmax attention.}
Beyond the scope of the existing works \citep{lee2023recasting} focusing on basic MCL formulation and setting, \dgv{we conduct a systematic empirical investigation of constant-state learners for MCL along three aspects: (i) \emph{effectiveness} across task types, spanning general and fine-grained classification and regression episodes; (ii) \emph{generalization} beyond the meta-training distribution, to longer episodes, unseen domains, and noisy inputs; and (iii) \emph{model design}, comparing architectures, components, and variants (\eg, with mixture-of-experts)}. 
\dgv{Our experiments show that constant-state learners can be meta-learned to perform well across most MCL scenarios: Mamba significantly outperforms kernel-based linear-attention models, and matches or surpasses Transformers with fewer parameters, constant memory, and less computation.} Notably, in some challenging settings with strong global structures (\eg, fine-grained) and scenarios with domain shifts or long sequences, Mamba demonstrates promising reliability, generalization, and robustness, highlighting its potential for efficient CL and adaptation. \dgv{Our main contributions are:
\begin{itemize}
\vspace{-3pt}
\item We formulate constant-state sequence models, in particular the selective SSM (e.g., Mamba), as meta-learned continual learners for MCL, whose bounded state matches CL's requirement of not retaining all seen samples.
\vspace{-3pt}
\item We meta-train the constant-state learners with a proposed selectivity regularization to make them work well: it supervises the implicit associations within the compressed state, stabilizing the challenging meta-training, and applies across different constant-state models with no cost at inference.
\vspace{-3pt}
\item We conduct a systematic empirical study of these learners across effectiveness, generalization, and model design, characterizing their behaviors in diverse MCL scenarios rather than claiming superiority over attention.
\end{itemize}}

\section{Related Work}
\textbf{Continual learning} aims to address catastrophic forgetting when training models across sequential tasks~\citep{de2021continual,wang2024comprehensive}. Replay-based methods retain or generate past samples~\citep{rebuffi2017icarl,lopez2017gradient,chaudhry2019tiny,shin2017continual,rostami2019complementary,riemer2019scalable}; regularization-based methods penalize changes to important parameters or apply distillation~\citep{kirkpatrick2017overcoming,zenke2017continual,li2017learning,aljundi2018memory,zhang2020class}; architecture-based methods assign task-specific parameters via masking or dynamic expansion~\citep{yoon2017lifelong,li2019learn,yan2021dynamically,ye2023self,wang2024self,lu2024adaptive}.

\noindent\textbf{Meta-continual learning} (MCL) meta-learns a continual learner over multiple CL episodes, structured into meta-training and meta-testing sets~\citep{son2024meta,lee2023recasting,bornschein2024transformers}, going beyond broader combinations of meta-learning and CL~\citep{riemer2018learning,beaulieu2020learning,gupta2020look,wu2024meta,mai2021supervised}: \cite{lee2023recasting} frames MCL as sequence modeling, while OML~\citep{javed2019meta} performs SGD-based adaptation atop a meta-learned encoder. \rev{The sequence-modeling formulation is closely related to in-context learning (ICL), where sequence models adapt to new tasks from the context without parameter updates~\citep{brown2020language,garg2022can,min2021metaicl}, with recent works studying the ICL abilities of attention-free and test-time-memorizing models~\citep{park2024can,sun2024learning,behrouz2024titans}; MCL differs in that the context is a non-stationary stream under CL's memory constraints.}
\begin{figure}[!t]
    \begin{center}
    \includegraphics[width=0.93\linewidth]{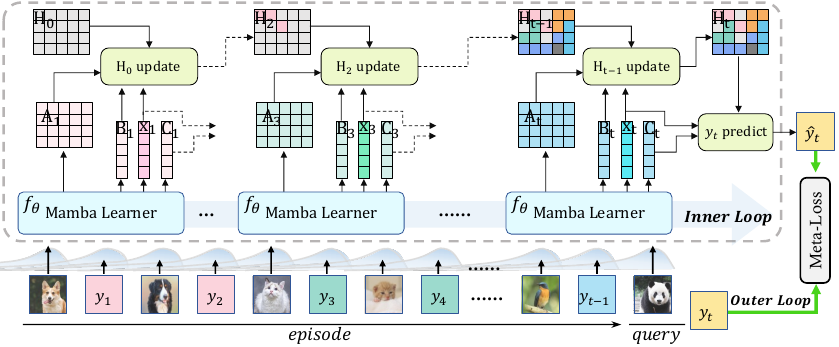}
    \caption{The overall framework of our proposed methods. We meta-train a Mamba Learner $f_{\theta}()$ to perform meta-continual learning (MCL) by processing an online data stream containing paired $(\rvx, y)$ examples. Meta-learning of this continual learner is conducted across multiple CL episodes. The model produces predictions by relying on the retained hidden state. Here, we demonstrate how the Mamba learner recurrently processes input data at steps $0$, $2$, and $t-1$, respectively.}
        \vspace{-0.4cm}
    \label{fig:mambacl}
    \end{center}
\end{figure}

\par
\noindent\textbf{Transformer} architectures~\citep{vaswani2017attention}, especially causal decoder-only variants~\citep{brown2020language}, excel at sequence modeling, but their softmax attention incurs quadratic complexity with sequence length. Linear attention methods replace the exponential similarity with kernelized dot products, enabling RNN-like linear-time modeling~\citep{katharopoulos2020transformers,choromanski2020rethinking,tay2020efficient}.

\noindent\textbf{State Space Models} (SSMs), rooted in classical control theory~\citep{kalman1960new}, enable efficient linear-time sequence modeling~\citep{gu2021efficiently,gu2021combining}. Mamba introduces input-dependent (selective) parameters with a hardware-efficient implementation~\citep{gu2023mamba,dao2024transformers}, proving effective across language and vision tasks~\citep{zhu2024vision,han2024demystify,lieber2024jamba}. \rev{A broader family of modern recurrent models with input-dependent state updates, such as RWKV~\citep{peng2023rwkv,peng2025rwkv} and Longhorn~\citep{liu2025longhorn}, similarly achieves competitive performance with constant-size states.}

\section{Problem Formulation and Methodology}
\rev{In \textbf{Continual Learning} (CL), given a non-stationary data stream $\gD_T^\text{train} = \{(\rvx_t, y_t)\}_{t=1}^T$ with $\rvx_t\in \gX_t$ and $y_t\in \gY_t$, a predictive model $P_{\phi_t}(y|\rvx)$ with parameters $\phi_t$ at step $t$ is trained on the stream and evaluated on a testing set $\gD_T^\text{test}=\{ (\rvx_n, y_n) \}_{n=1}^N$ following the distribution of the training stream.}
A conventional continual learner is manually crafted to continually update model parameters $\phi_t$ through optimization. 
The data stream \(\gD_T\) typically comprises samples from different tasks or distributions, often piecewise stationary within a task interval. In online CL, each sample is seen only once, whereas in offline CL, task-specific samples can be accessed for multiple rounds. This work primarily focuses on the online CL setting.

\rev{To achieve} efficient CL, \textbf{Meta-Continual Learning} (MCL) \citep{lee2023recasting,son2024meta} meta-learns a parameterized continual learner that updates a predictive model (in $\gX \rightarrow \gY$) from a streaming data source (in $\gX\times \gY$). Given a learned model from a sequence $\gD_t^{\text{train}}$ in MCL, prediction can be formulated as \( P_{\theta}(y_\text{test}|\rvx_\text{test}, \gD_t^\text{train}) \). MCL can be seen as training a \emph{sequence prediction} (SP) model \rev{$f_\theta():(\gX\times \gY)^{*}\times \gX \rightarrow \gY$ over sequences of sample pairs}, where $f_\theta()$ continually processes streaming data and performs prediction for test samples via \rev{$\hat{y}_\text{test} = f_\theta( (\gD_t^\text{train}, \rvx_\text{test}) )$}. The learner updates its hidden states dynamically, reflecting the continual learning process.
Through multiple episodes of $(\gD^\text{train}, \gD^\text{test})$, the learner’s parameters $\theta$ are optimized via meta-learning to improve performance across all \rev{episodes}. The targets $y$ in different episodes are independent, not sharing semantic meaning.
This work focuses on MCL using a parameterized SP model, \rev{in contrast to gradient-based meta-learning approaches} \citep{finn2017model,javed2019meta}.

\subsection{Preliminaries: Transformers and SSMs}

\textbf{Transformers} produce next-token predictions in sequence relying on a self-attention mechanism \citep{vaswani2017attention}. \rev{Given a sequence of $N$ vectors $\rmZ\in \sR^{N\times M}$, softmax attention projects the tokens into queries, keys, and values, \ie, $\rmQ\!=\rmZ\rmW_Q$, $\rmK\!=\rmZ\rmW_K$, $\rmV\!=\rmZ\rmW_V$, and produces the outputs as $\rvu_t=\sum_{j=1}^{N} \mathrm{softmax}_j\big(\rmQ_t\rmK_j^\top/\sqrt{d}\big)\rmV_j$, where $\rmQ_t$, $\rmK_j$, and $\rmV_j$ denote the indexed row vectors and $d$ is the query/key dimension.}
Each input token generates a key-value pair, leading to a linear increase in key-value cache size.
Softmax attention measures the similarities between the query-key pairs, leading to $\gO(N^2)$ complexity.

\par
\noindent\textbf{Linear Transformer} \citep{katharopoulos2020transformers} reduces the complexity relying on a \emph{linear attention} method\rev{, which applies a feature map $\phi(\cdot)$ corresponding to a kernel to the queries and keys, \ie, $\rmQ\!=\!\phi(\rmZ\rmW_Q)$, $\rmK\!=\!\phi(\rmZ\rmW_K)$ with $\phi(\rvx)=\mathrm{elu}(\rvx)+1$, replacing the softmax with linear operations of $\gO(N)$ complexity; \textbf{Performer}~\citep{choromanski2020rethinking} instead employs random-feature maps.}

\par
\rev{In the autoregressive (causal) setting, where the computation of $\rvu_t$ only sees preceding tokens $j\leq t$, linear attention can be computed recurrently:}
{\small
\begin{equation} \label{eq:linear_att_causal}
\rvu_t=\frac{\rmQ_t\rmS_t}{\rmQ_t\rmG_t}, ~~
\rmS_t=\rmS_{t-1}+{\rmK_t^{\top}\rmV_t}, ~~\rmG_t=\rmG_{t-1}+{\rmK_t^{\top}},
\end{equation}}
\rev{where the cumulatively updated $\rmS_t=\sum_{j=1}^{t}\rmK_j^{\top}\rmV_j$ and $\rmG_t=\sum_{j=1}^{t}\rmK_j^{\top}$ serve as constant-size internal hidden states.}

\par
\noindent\textbf{State space sequence models (SSMs)} \citep{gu2023mamba,gu2021efficiently,dao2024transformers} \rev{map an input sequence $z_t\in\sR$ to outputs $u_t\in\sR$ through a hidden state $\rvh_t\in \sR^{C\times 1}$, formulated with parameters $\rmA\in \sR^{C\times C}$, $\rmB\in\sR^{C\times 1}$, $\rmC\in\sR^{1\times C}$, and $D\in \sR$ as $\rvh_t = \rmA \rvh_{t-1} + \rmB z_t,~ u_t = \rmC \rvh_t + D z_t$, where the discrete $\rmA$ and $\rmB$, transformed from continuous counterparts via a timescale parameter $\Delta$, govern the hidden-state update.}

\subsection{SSM and Mamba for Meta-Continual Learning}

\par
\textbf{Selective SSM \& Mamba in MCL.} The dynamics of the basic SSM are time-invariant, limiting \rev{its} ability to model complex sequences.
\rev{Mamba \citep{gu2023mamba} instead adopts a \emph{selective} SSM, generating input-dependent parameters for step-sensitive selection:}
{\small
\begin{equation}
\rvh_{t} = \rmA_t\rvh_{t-1} + \rmB_t z_t,~~ u_t=\rmC_t\rvh_t+ D z_t,
\label{eq:slec_ssm}
\end{equation}}
\rev{where $\rmA_t$, $\rmB_t$, and $\rmC_t$ are generated from the input token at step $t$, controlling how the fixed-size hidden state is retained ($\rmA_t$), written ($\rmB_t$), and read ($\rmC_t$) at each step.}

\par
\rev{To handle vector tokens $\rvz_t\in\sR^M$ in MCL, Mamba applies the selective SSM to each dimension/channel independently:}
{\small
\begin{equation}
\rmH_{t} = [\rmA_{t,i}\rvh_{t-1,i} + \rmB_{t,i} \rvz_{t,i}]_{i=1}^M,~\rvu_t=\rmC_t\rmH_t+ \rmD \odot \rvz_t,
\label{eq:slec_ssm_vec}
\end{equation}}
where $\rmH_t\in \sR^{C\times M}$ is the concatenated hidden state across all $M$ input dimensions, $\rmC_t\in \sR^{1\times C}$, $\rmD\in \sR^{1\times M}$, and $\rvu_t\in \sR^{1\times M}$. \rev{In Mamba-2, $\rmA_{t,i}$ is tied within each head (a per-head scalar), and $\rmB_t$ and $\rmC_t$ are shared across channels~\citep{dao2024transformers}.}
The Mamba block used in our work applies a 1-D convolution on the input tokens and then projects the representations to obtain input-dependent SSM parameters \citep{dao2024transformers}\rev{ (see the appendix for an illustration); multiple blocks are stacked homogeneously. Unless otherwise specified, Mamba refers to Mamba-2 throughout the paper.}


\subsubsection{MCL with Mamba as a Continual Learner} 
\rev{We meta-learn the Mamba model $f_\theta()$ as a continual learner over multiple CL episodes $(\gD^\text{train}, \gD^\text{test})\sim P_{(\gX, \gY)}$, where learning and prediction within an episode are performed as sequence prediction through the recurrently retained hidden state. Each episode is framed as next-token prediction over the sequence $(\rvx^\text{train}_{1}, y^\text{train}_{1}, ..., \rvx^\text{train}_{T}, y^\text{train}_{T}, \rvx^\text{test})\rightarrow y^\text{test}$; for instance, in class-incremental learning, all testing classes have appeared in the preceding stream. The learner's parameters $\theta$ are meta-optimized across sampled episodes:}
{\small
\begin{equation} \label{eq:metacl_obj}
\begin{aligned}
    \min_\theta  ~~ & \E_{ (\gD^\text{train}, \gD^\text{test})\sim P_{(\gX, \gY)} } \\
    & \sum\nolimits_{ (\rvx^\text{test}, y^\text{test})\in \gD^\text{test} }
         \ell\Big( f_\theta\big( (\gD^\text{train},\rvx^\text{test}) \big) ,\, y^\text{test} \Big),
\end{aligned}
\end{equation}}
where $\ell(\cdot, \cdot)$ denotes the proper loss function for different tasks, \eg, classification or regression. \rev{Pseudocode of the meta-training and meta-testing procedures is provided in the appendix.}

\par
\rev{On the data stream, the meta-learned Mamba $f_\theta()$ learns to associate inputs with their targets along the stream, recurrently updating the hidden state $\rmH_t$ for prediction, as shown in Fig.~\ref{fig:mambacl}. This efficient online CL process compresses knowledge from the stream into the state in a content-aware manner. To explore extensions of MambaCL, we also incorporate mixture-of-experts (MoE) into the Mamba model \citep{fedus2022switch,pioro2024moe} for learning and mixing multiple learners (see the ablation studies).}

\noindent\textbf{Target token embeddings.} 
\rev{Since targets $y$ are symbolic, \ie, consistent within an episode but carrying no shared meaning across episodes, the learner must handle arbitrary episodes and generalize across domains. Instead of pre-defining a small, fixed set of candidate targets (\eg, classes) with a restricted prediction head, we represent targets by token embeddings from a large universal vocabulary \citep{lee2023recasting}, inspired by tokenization in LMs \citep{sennrich2015neural,devlin2019bert}: each episode randomly picks a subset of unique codes to indicate its classes, and at inference the model predicts the next token over the full vocabulary.}
\rev{Beyond the standard setting where meta-training and meta-testing share the same number of classes \citep{lee2023recasting}, we also analyze generalization across differing configurations in the experiments.}

\subsubsection{Selectivity Regularization for Meta-Training}
\label{sec:regularization}
\rev{Meta-learning a continual learner to associate inputs and targets over a data stream is challenging for Transformers and attention-free models \citep{lee2023recasting}: the supervision signal depends on long-range interactions across the stream, making meta-training slow to converge. We thus provide additional guidance that strengthens the associations, known in meta-training, between each input token and its correlated preceding tokens. While such guidance is straightforward for Transformers with explicit intra-sequence attention \citep{lee2023recasting}, it is non-trivial for SSM-based Mamba.}

\rev{During training, consider the input token of the $(t{+}1)$-th sample $(\rvx_{t+1}, y_{t+1})$, arriving at step $2t+1$ after the $2t$ tokens of the preceding $t$ samples. Its ground-truth association with the preceding tokens is $\rvp_{2t+1}=[\mathds{1}_{y_{t+1}}(y_1), \mathds{1}_{y_{t+1}}(y_1), ..., \mathds{1}_{y_{t+1}}(y_t), \mathds{1}_{y_{t+1}}(y_t)]\in \{0,1\}^{2t}$, where the indicator $\mathds{1}_y(y')$ equals $1$ if $y=y'$ and $0$ otherwise, and each entry repeats twice since every sample contributes two tokens ($\rvx_j$ and $y_j$). We encourage the meta-learned learner to internalize this association pattern so that it can be reused in CL (\ie, meta-testing).}

\par
\rev{Transformers maintain the key-value pairs of all seen tokens as their state and retrieve information through explicit attention. For the token at step $2t+1$, the attention pattern over previous tokens is $\rvq_{2t+1}^{\text{Trans}} = [\rmQ_{2t+1}\rmK_j^\top]_{j=1}^{2t}\in \sR_{\geq 0}^{2t}$ (normalization omitted for simplicity), so the guidance can be directly imposed by encouraging $\rvq_{2t+1}^{\text{Trans}}$ to match $\rvp_{2t+1}$.}

\par
\rev{Mamba and other attention-free models instead compress the history into a hidden state (Eqs.~(\ref{eq:linear_att_causal}) and~(\ref{eq:slec_ssm})) and produce no explicit attention weights. We therefore construct their association patterns through the connection to attention. For the Linear Transformer, $\rmQ$ and $\rmK$ retain the same query/key roles as in softmax attention, so storing the intermediate $\rmK_j$ \emph{only during training} yields $\rvq_{2t+1}^{\text{LNTrans}} = [\rmQ_{2t+1}\rmK_j^\top]_{j=1}^{2t}$. For Mamba, the duality between the selective SSM (Eqs.~(\ref{eq:slec_ssm}) and~(\ref{eq:slec_ssm_vec})) and linear attention~\citep{dao2024transformers} identifies the selective parameters $\rmC_t$ and $\rmB_t$ with the query and key embeddings $\rmQ_t$ and $\rmK_t$, giving the associative indicators $\rvq_{2t+1}^{\text{Mamba}} = [\rmC_{2t+1}\rmB_j^\top]_{j=1}^{2t}$. For each input token in the stream, we then penalize the divergence between the model's association pattern and the ground truth, $\ell_\text{slct} = \text{KL}(\rvp_{2t+1}, \rvq^{*}_{2t+1})$, where $*$ denotes the specific model. The loss is accumulated over the stream, weighted by a scalar $\lambda$, and optimized together with the MCL objective in Eq.~(\ref{eq:metacl_obj}); it is applied to MambaCL and all compared sequence models to enhance meta-training stability and convergence. The stored intermediate components are needed only during meta-training and incur no inference cost.}
Further details are provided in \rev{the ablation studies and the appendix}, where we report the
$\lambda$ values used in our selectivity regularization loss and present the training loss curves with and without selectivity regularization for different models.
\section{Experiments and Analyses}

We evaluate different models for MCL and analyze their behavior across various scenarios. Our meta-learning episodes primarily involve classification tasks but also include regression. Unlike previous studies focused on basic settings, we conduct experiments in both standard MCL setups and more diverse configurations.

\noindent \textbf{Experimental setup.} 
\rev{To obtain MCL tasks, given a classification dataset, we first divide it into multiple tasks, each representing a distinct set of classes. These tasks are split into two non-overlapping sets for meta-training and meta-testing. CL episodes in both sets are constructed identically: each episode randomly selects $K$ distinct tasks, with $K$ set to 20 by default. We also explore scenarios with varying $K$ values.}
By default, each task in both training and testing sequences contains five samples (5-shot). We also explore varying shot numbers to further evaluate adaptability and learning efficiency. For implementation details, please refer to \rev{the appendix}.

\noindent \textbf{Datasets.}
\rev{We conduct experiments on various datasets: \textbf{general image classification} with CIFAR-100~\citep{krizhevsky2009learning}, ImageNet-1K~\citep{russakovsky2015imagenet}, ImageNet-R~\citep{hendrycks2021many}, MS-Celeb-1M (Celeb)~\citep{guo2016ms}, CASIA Chinese handwriting (Casia)~\citep{liu2011casia}, and Omniglot~\citep{lake2015human}; \textbf{fine-grained recognition} with CUB-200~\citep{wah2011caltech}, Stanford Dogs~\citep{dataset2011novel}, Stanford Cars~\citep{krause20133d}, and FGVC-Aircraft (Aircraft)~\citep{maji2013fine}; \textbf{large domain shift} with DomainNet~\citep{peng2019moment}; and \textbf{regression} with sine wave reconstruction (Sine), image rotation prediction (Rotation), and image completion (Completion).}
Additional details are provided in \rev{the appendix}.

\begin{table}[t]
\caption{\rev{Classification accuracy (\%) on 20-task 5-shot MCL with pre-trained representations}, on general image classification tasks (left) and fine-grained recognition tasks (right). \highred{Best} and \highblue{2nd best} results are in \highred{red} and \highblue{blue}, respectively.}
\label{tab:clip}
\begin{center}
\small
    \setlength{\tabcolsep}{4pt}
    \resizebox{1.0\linewidth}{!}
    {
    \begin{tabular}{lcccccccccc}
    \toprule
        \multirow{2}{*}{Method} &\multicolumn{6}{c}{General Image Classification} &\multicolumn{4}{c}{Fine-grained Recognition} \\
        \cmidrule(lr){2-7}\cmidrule(lr){8-11}
        &CIFAR-100 &ImageNet-1K &ImageNet-R &Celeb  &Casia &Omniglot &CUB-200 &Dogs &Cars &Aircraft \\
    \midrule
        OML        &\highblue{$64.4^{\pm0.4}$} &$90.5^{\pm0.3}$ &\highblue{$67.5^{\pm0.3}$} &$72.8^{\pm0.1}$ &$81.5^{\pm0.5}$ &$90.4^{\pm0.2}$ &$78.7^{\pm0.6}$ &$72.4^{\pm0.5}$ &$83.6^{\pm0.7}$ &$49.5^{\pm0.2}$ \\
        Transformer&$62.7^{\pm0.7}$ &\highblue{$93.5^{\pm0.1}$} &$63.6^{\pm0.2}$ &\highred{$78.4^{\pm0.1}$} &\highred{$93.8^{\pm0.2}$} &\highblue{$94.4^{\pm0.2}$} &\highblue{$81.4^{\pm0.4}$} &\highblue{$77.5^{\pm0.6}$} &\highblue{$87.0^{\pm0.3}$} &\highblue{$53.9^{\pm0.7}$} \\
        Linear TF  &$54.3^{\pm0.7}$ &$89.1^{\pm0.2}$ &$55.7^{\pm0.3}$ &$76.5^{\pm0.2}$ &$90.9^{\pm0.4}$ &$86.5^{\pm0.5}$ &$69.7^{\pm0.7}$ &$69.7^{\pm0.7}$ &$76.6^{\pm0.8}$ &$49.0^{\pm0.7}$ \\
        Performer  &$53.4^{\pm0.3}$ &$90.8^{\pm0.5}$ &$52.8^{\pm0.9}$ &$76.8^{\pm0.1}$ &$93.0^{\pm0.3}$ &$89.3^{\pm0.3}$ &$69.2^{\pm0.8}$ &$69.4^{\pm0.4}$ &$73.9^{\pm0.8}$ &$48.6^{\pm0.6}$ \\
        Mamba      &\highred{$67.1^{\pm0.4}$} &\highred{$93.6^{\pm0.2}$} &\highred{$69.7^{\pm0.4}$} &\highblue{$77.0^{\pm0.1}$} &\highblue{$93.1^{\pm0.2}$} &\highred{$95.9^{\pm0.2}$} &\highred{$83.0^{\pm0.4}$} &\highred{$79.2^{\pm0.5}$} &\highred{$88.3^{\pm0.4}$} &\highred{$55.3^{\pm0.6}$} \\
    \bottomrule
    \end{tabular}
    }
\end{center}
\end{table}

\subsection{{Experimental Results and Analyses}}
\label{sec:main_experiment}
We evaluate several models, including OML~\citep{javed2019meta}, Vanilla Transformer~\citep{vaswani2017attention}, Linear Transformer~\citep{katharopoulos2020transformers}, Performer~\citep{choromanski2020rethinking}, and MambaCL. \rev{OML serves as an SGD-based MCL baseline with a two-layer MLP atop a meta-learned encoder, while others are meta-learned sequence-prediction learners. All sequence models use 4 layers with hidden dimension 512, and MambaCL has significantly fewer parameters under the same depth and width.}

\begin{figure}[!t]
\begin{center}
    \includegraphics[width=1.0\linewidth]{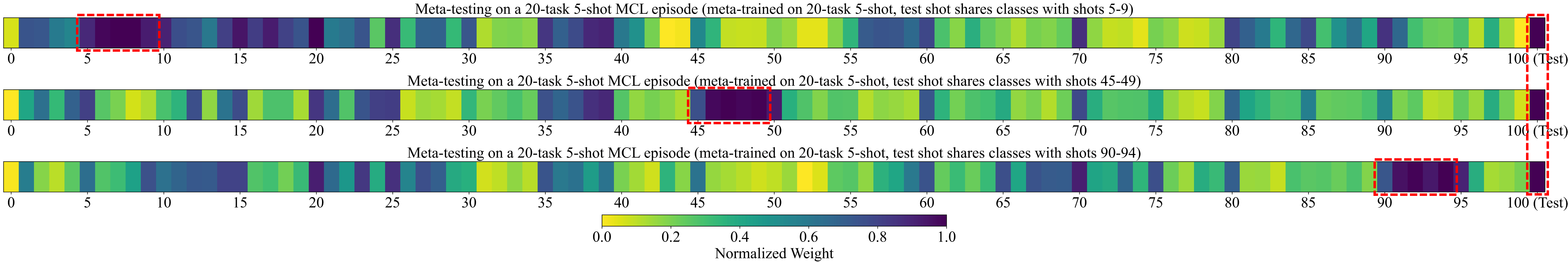}
    \caption{Final-layer associations in a 20-task 5-shot meta-testing episode of MambaCL (meta-trained on 20-task 5-shot MCL). 
    The three visualizations share the same training episode (shots $0^{th}\!\!-\!99^{th}$), but differ in their test queries at the $100^{th}$ shot. Specifically, these test queries come from the {$1^{st}$}, {$9^{th}$}, and {$18^{th}$} tasks in the training episode, respectively, and therefore each query shares the same class label with the $5^{th}\!\!-\!9^{th}$, $45^{th}\!\!-\!49^{th}$, and $90^{th}\!\!-\!94^{th}$ training shots within the episode.
    The \textcolor{red}{red box} highlights each query and its corresponding training shots within the episode. See \rev{the appendix} for additional visualizations comparing Transformer attention and Mamba's associative selectivity.}
    \vspace{-0.3cm}
    \label{fig:mambacl_vis}
    \end{center}
\end{figure}

\noindent\textbf{General image classification tasks.}
Table~\ref{tab:clip} \rev{(left)} provides comparative analyses of the performance of various architectures across a range of general image classification tasks, utilizing image representations derived from a pre-trained encoder. \rev{Results of training from scratch are provided in the appendix, where Mamba performs best or second best across datasets, and all methods suffer from substantial meta-overfitting on CIFAR-100 (meta-testing accuracy below $19\%$), possibly due to its lower task diversity.} In Table~\ref{tab:clip}, continual learners based on pre-trained models show \rev{trends consistent with the from-scratch results} on CIFAR-100 and ImageNet-R, with Mamba outperforming other methods, highlighting its robustness against overfitting. On larger datasets like ImageNet-1K, Casia, and Celeb, Mamba matches or exceeds \rev{Transformer} performance. \rev{We default to pre-trained image representations for the remaining experiments.} To better understand its behavior, we visualize the associative weights of our meta-trained Mamba Learner in Fig.~\ref{fig:mambacl_vis}. For further details on the visualization analysis of Transformers and Mamba, please refer to \rev{the appendix}.

\noindent\textbf{Fine-grained recognition tasks.}
Table~\ref{tab:clip} \rev{(right)} presents a performance comparison of different architectures on fine-grained recognition datasets.
In fine-grained datasets characterized by subtle inter-class differences (\eg, the CUB-200 dataset, which contains 200 bird subcategories), models need to integrate global information throughout the training episode to identify these nuances effectively.
Mamba outperforms other models across these datasets, potentially attributable to its robustness to capture subtle inter-class distinctions.
\rev{Notably, comparing the two halves of Table~\ref{tab:clip}: while Mamba and Transformers trade the lead on general tasks, Mamba leads on all four fine-grained datasets, where integrating global context within the episode matters most.}

\begin{table}[t]
\caption{Classification accuracy (\%) and regression errors \rev{on} 100-task 5-shot MCL.}
\label{tab:scratch100}
\begin{center}
  \small
  \resizebox{0.7\linewidth}{!}
{
  \begin{tabular}{lccccc}
    \toprule
        Method & Casia & Celeb & Sine & Rotation & Completion\\
    \midrule
         OML & $93.2^{\pm0.9}$ & $45.5^{\pm0.2}$ & $0.0498^{\pm0.0004}$ & $0.524^{\pm0.087}$ & $0.1087^{\pm0.0001}$\\
        Transformer & \highblue{$99.0^{\pm0.0}$} & \highred{${60.5}^{\pm0.1}$} & \highred{${0.0031}^{\pm0.0002}$} & \highblue{${0.031}^{\pm0.001}$ }& \highblue{${0.0989}^{\pm0.0001}$}\\
        Linear TF & $97.7^{\pm0.1}$ & $54.7^{\pm0.1}$ & ${0.0139}^{\pm0.0003}$ & ${0.047}^{\pm0.002}$ & ${0.1084}^{\pm0.0001}$\\
        Mamba  &\highred{$99.1^{\pm0.1}$} &\highblue{$59.9^{\pm0.1}$}& \highblue{$0.0054^{\pm0.0001}$} &  \highred{$0.025^{\pm0.001}$ } &  \highred{$0.0895^{\pm0.0001}$}\\
    \bottomrule
  \end{tabular}
  }
\end{center}
\end{table}

\noindent\textbf{Training on Longer Episodes.} \label{sec:experiment_regression}
We conducted experiments to meta-train the models on longer episodes across both classification and regression tasks. Table~\ref{tab:scratch100} demonstrates that Mamba continues to perform comparably to Transformer, and significantly outperforms SGD-based approaches (OML).

\subsection{Generalization Analyses}

\label{sec:generalization}
\rev{A meta-learned learner should generalize to scenarios unseen during meta-training. To assess this, we analyze Transformers and Mamba under sequence lengths beyond those seen in meta-training, significant domain shifts, and noisy inputs during meta-testing.} Additionally, to understand the behavior of these models, we visualize the attention weights of Transformers and the associative weights of Mamba, illustrating their attention mechanisms and selectivity patterns in \rev{the appendix}.

\noindent\textbf{Generalization to varying sequence length.} 
To address continual learning episodes of indefinite length effectively, the learning algorithm should demonstrate the capability to generalize beyond the sequence lengths encountered during meta-training. 
We conducted length generalization experiments on ImageNet-1K, training vanilla and linear Transformers, and Mamba on 20-task 5-shot MCL, each with a vocabulary of 200 tokens. The length of a continual learning episode is calculated as $2\times \text{tasks} \times \text{shots} + 1$.

\begin{figure}[t]
  \centering
  \begin{minipage}[t]{0.57\linewidth}
    \centering
      \begin{minipage}[t]{0.3\linewidth}
          \centering
          \includegraphics[width=1\linewidth]{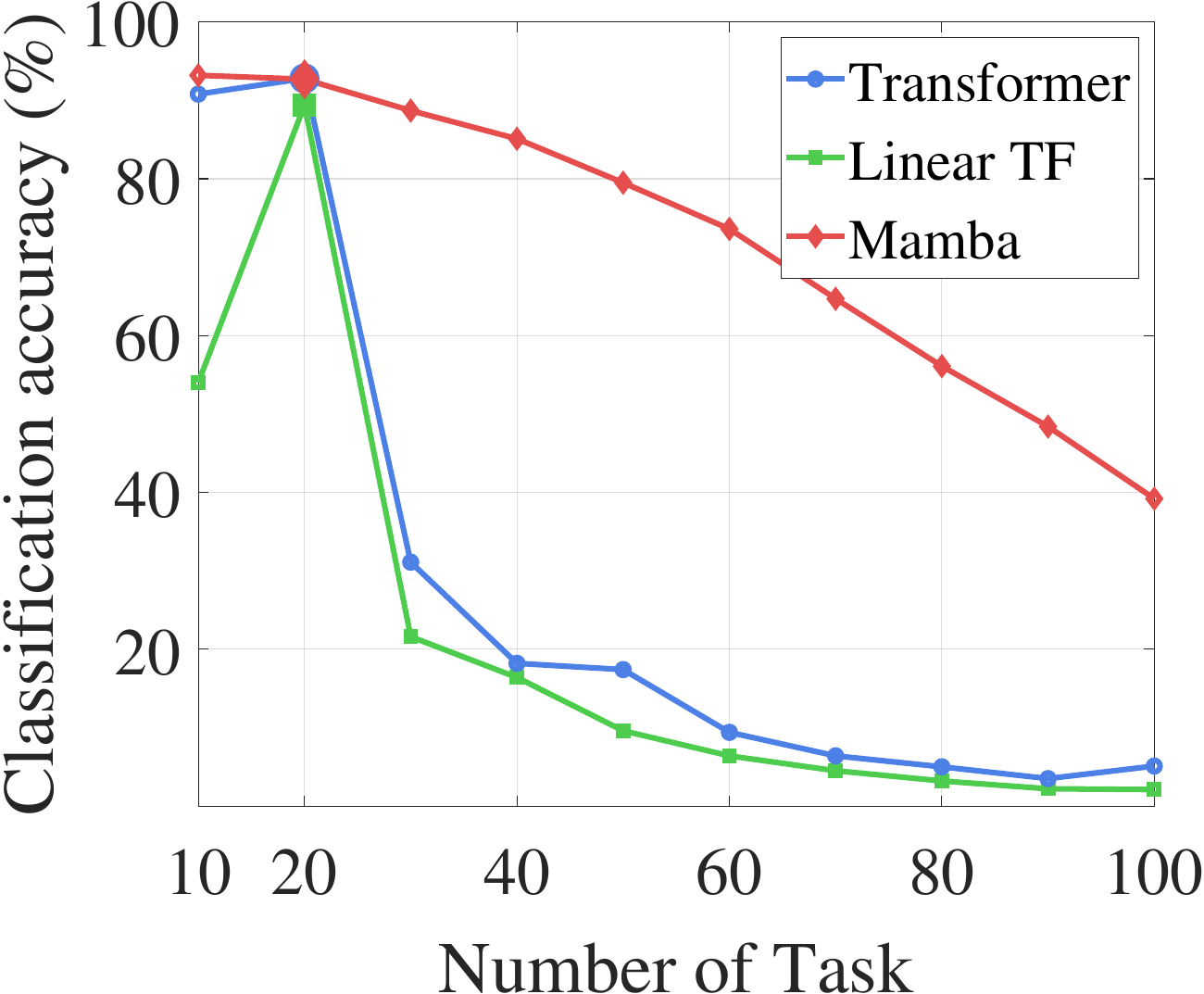}
          \subcaption{\small Task}
          \label{fig:generalization-task}
      \end{minipage}
      \hspace{1pt}
      \begin{minipage}[t]{0.3\linewidth}
          \centering
          \includegraphics[width=1\linewidth]{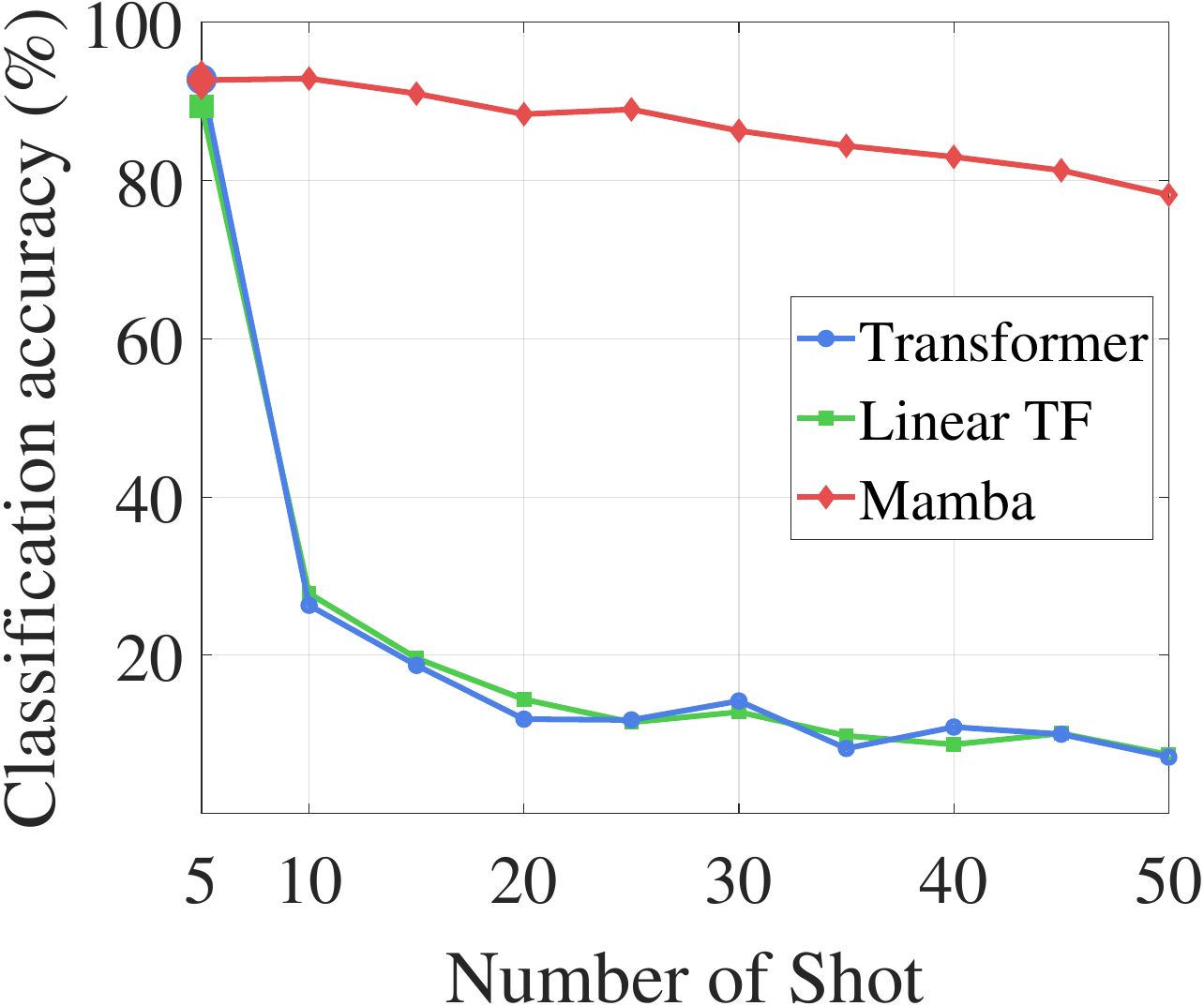}
          \subcaption{\small Shot}
          \label{fig:generalization-shot}
      \end{minipage}
      \hspace{1pt}
      \begin{minipage}[t]{0.3\linewidth}
          \centering
          \includegraphics[width=1\linewidth]{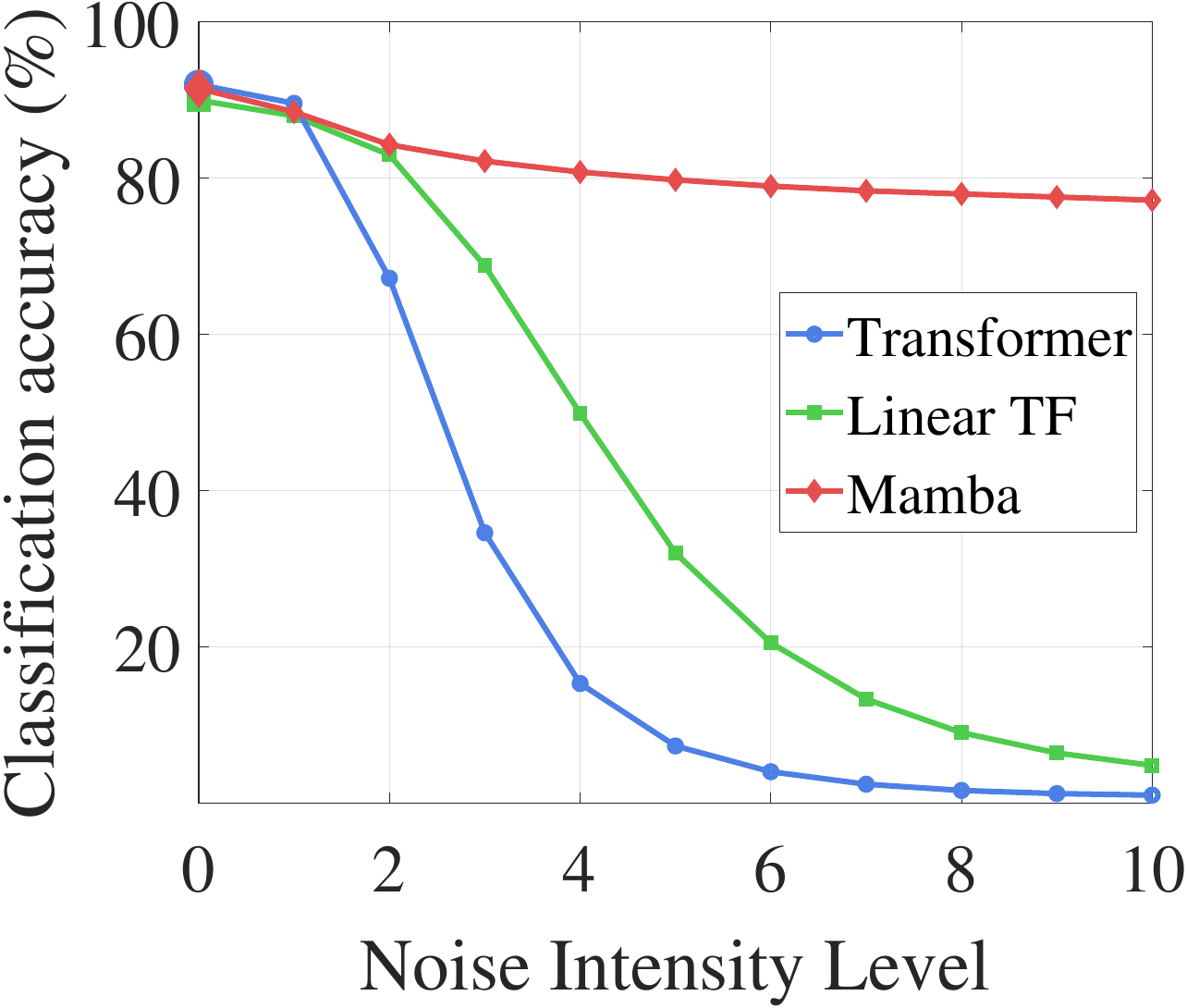}
          \subcaption{\small Noise}
          \label{fig:generalization-noise}
      \end{minipage}
      \caption{Generalization analysis on ImageNet-1K, with meta-training on 20 tasks (5-shot), meta-testing on: (a) varying tasks (5-shot), (b) varying shots (20-task), and (c) varying input noise levels (20-task 5-shot).}
      \label{fig:generalization}
  \end{minipage}
  \hfill
  \begin{minipage}[t]{0.39\linewidth}
    \centering
      \begin{minipage}[t]{0.48\linewidth}
          \centering
          \includegraphics[width=1\linewidth]{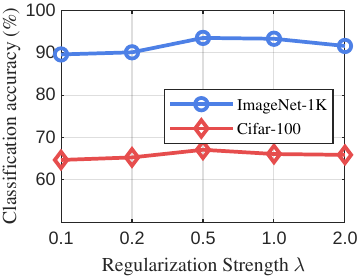}
          \subcaption{\small $\lambda$}
          \label{fig:lambda}
      \end{minipage}
      \hspace{1pt}
      \begin{minipage}[t]{0.48\linewidth}
          \centering
          \includegraphics[width=1\linewidth]{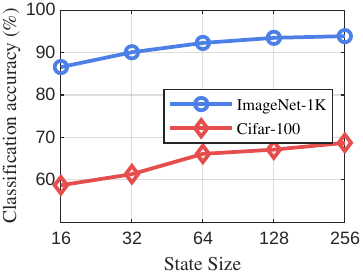}
          \subcaption{\small SSM state size}
          \label{fig:state}
      \end{minipage}
      \caption{Ablation studies: (a) varying $\lambda$ in training loss; (b) varying SSM state size.}
    \end{minipage}
\end{figure}

\noindent\textbf{\textit{(a) Meta-testing on varying numbers of tasks.}}\label{sec:experiment_task}
Fig.~\ref{fig:generalization-task} illustrates the performance of the three models, meta-trained on a 20-task 5-shot setup and evaluated during meta-testing across varying numbers of tasks while keeping a constant shot number of 5. Both the vanilla Transformer and Linear Transformer suffer significant performance degradation during meta-testing at untrained episode lengths, even in configurations such as the 10-task 5-shot setup. 
Mamba demonstrates superior performance in the 10-task setup compared to the meta-trained 20-task setup, and its performance degradation is relatively mild compared to Transformers as the number of tasks increases.

\noindent\textbf{\textit{(b) Meta-testing on varying numbers of shots.}}\label{sec:experiment_shot}
In Fig.~\ref{fig:generalization-shot}, we assess the performance of three models meta-trained in a 20-task 5-shot setup and evaluated during meta-testing across varying \rev{shot numbers} while keeping a constant task number of 20. Both the vanilla Transformer and linear Transformer suffer significant performance degradation, likely due to overfitting the 20-task 5-shot pattern. In contrast, Mamba exhibits only about a 10\% performance degradation when the number of shots in meta-testing increases to 50, which is ten times the length of the meta-training episodes. Fig.~\ref{fig:generalization-task} and~\ref{fig:generalization-shot} highlight Mamba's robustness in adapting to longer sequence lengths (length generalization).

\begin{table}[t]
\caption{\rev{Classification accuracy (\%) on the DomainNet dataset (20-task 5-shot MCL)}. Each column reports the domain held out for meta-testing, with the remaining five domains used for meta-training. (\textit{clp: clipart, inf: infograph, pnt: painting, qdr: quickdraw, rel: real, skt: sketch.})}
\label{tab:domain}

\begin{center}
\small
    \setlength{\tabcolsep}{3pt}
    \resizebox{0.8\linewidth}{!}
    {
    \begin{tabular}{lcccccc|c}
    \toprule
        Method &clp &inf &pnt &qdr &rel &skt &Avg \\
    \midrule
        Transformer&\highred{$91.8^{\pm0.1}$} &\highblue{$69.4^{\pm0.1}$} &\highred{$82.6^{\pm0.2}$} &\highblue{$50.2^{\pm0.6}$} &\highred{$93.8^{\pm0.1}$} &\highblue{$85.9^{\pm0.3}$} &\highblue{$79.0^{\pm0.3}$} \\
        Linear TF  &$91.0^{\pm0.0}$ &$66.2^{\pm0.8}$ &$80.6^{\pm0.8}$ &$30.7^{\pm1.4}$ &$92.9^{\pm0.1}$ &$85.5^{\pm0.1}$ &$74.5^{\pm0.5}$ \\
        Performer  &$91.3^{\pm0.2}$ &$66.4^{\pm0.6}$ &$81.3^{\pm0.2}$ &$39.4^{\pm1.7}$ &$92.8^{\pm0.1}$ &$84.8^{\pm0.5}$ &$76.0^{\pm0.6}$ \\
        Mamba      &\highblue{$91.7^{\pm0.2}$} &\highred{$70.2^{\pm0.2}$} &\highblue{$81.8^{\pm0.2}$} &\highred{$55.6^{\pm0.8}$} &\highblue{$93.0^{\pm0.1}$} &\highred{$87.2^{\pm0.3}$} &\highred{$79.9^{\pm0.3}$} \\
    \bottomrule
    \end{tabular}
    }
    \end{center}
    \vspace{-0.5cm}
\end{table}

\noindent\textbf{Results and analyses on larger domain shift.}
We investigate a larger domain shift scenario using the DomainNet dataset, which includes six distinct domains, to further assess model generalization to unseen input distributions, a setting that mirrors real-world conditions. One domain is held out for meta-testing, while the others are used for meta-training. The experimental results are detailed in Table~\ref{tab:domain}.
Overall, these models demonstrate the capability to handle large domain shift scenarios effectively. Mamba performs on par with or surpasses Transformers across diverse target domains, benefiting from potentially enhanced generalization capabilities associated with its smaller size and reduced susceptibility to overfitting. \rev{Vanilla Transformers perform well when the targets are real images or paintings.} Mamba excels in the quickdraw domain, which presents greater variances from other domains. This performance advantage may be ascribed to Mamba’s robustness in handling inputs that deviate significantly from the training distribution.

\noindent\textbf{Sensitivity to the noisy inputs.}
To assess the sensitivity of different models to noisy inputs, we conduct experiments on meta-trained 20-task 5-shot MCL models using the ImageNet-1K dataset. \rev{During each meta-testing episode, we add Gaussian noise with mean $0$ and standard deviation $\sigma\in[0, 10]$ to the input embeddings of \emph{five} randomly selected samples.}
As illustrated in Fig.~\ref{fig:generalization-noise}, the vanilla Transformer and Linear Transformer suffer significant performance degradation. In contrast, Mamba demonstrates robust performance, effectively handling inputs with increased \rev{noise levels}.

\subsection{Ablation Studies}

\begin{table}[t]
\begin{minipage}[t]{0.41\linewidth}
\caption{Different Mamba arch. on 20-task 5-shot MCL.}
\label{tab:ab}
\begin{center}
\small
\setlength{\tabcolsep}{3pt}
\resizebox{0.95\linewidth}{!}
{
\begin{tabular}{lcc}
\toprule
Method &CIFAR-100 &ImageNet-1K \\
\midrule
Transformer&{$62.7^{\pm0.7}$} &{$93.5^{\pm0.1}$}  \\
Linear TF  &$54.3^{\pm0.7}$ &$89.1^{\pm0.2}$  \\
\midrule
Mamba-1   &$59.7^{\pm0.5}$ &$90.1^{\pm0.3}$ \\
MambaFormer  &$62.4^{\pm0.6}$ &$92.7^{\pm0.1}$ \\
Mamba-2      &{$67.1^{\pm0.4}$} &{$93.6^{\pm0.2}$}    \\
Mamba+MoE   &{$68.9^{\pm0.2}$} &{$94.0^{\pm0.2}$}\\
\bottomrule
\end{tabular}
}
\end{center}
\end{minipage}
\hfill
\begin{minipage}[t]{0.56\linewidth}
\caption{\rev{Ablations of Mamba (\ie, Mamba-2, consisting of the selective SSM (S6), short convolution, and gating) on 20-task 5-shot MCL. Without the selective SSM, the model fails to converge, yielding random-guess accuracy.}}
\label{tab:component}
\begin{center}
\small
    \setlength{\tabcolsep}{2pt}
\resizebox{0.78\linewidth}{!}
{
\begin{tabular}{ccccc}

\toprule
SSM (S6) &Conv &Gating &CIFAR-100 &ImageNet-1K \\
\midrule
\checkmark &           &           &$54.2$ &$90.1$ \\
\checkmark &           &\checkmark &$55.3$ &$91.2$ \\
\checkmark &\checkmark &           &$65.6$ &$92.7$ \\
\checkmark &\checkmark &\checkmark &$\mathbf{67.1}$ &$\mathbf{93.6}$ \\
           &\checkmark &\checkmark &$5.0$ (random) &$5.0$ (random) \\
\bottomrule
\end{tabular}
}
\end{center}
\end{minipage}
\end{table}

\textbf{Hyper-parameter of selectivity regularization loss.}
\label{sec:main_regularization_ablation}
We conducted ablation studies to assess the impact of the selectivity regularization weight $\lambda$ on our Mamba model’s performance, varying it across ${0.1, 0.2, 0.5, 1.0, 2.0}$ (Fig.~\ref{fig:lambda}). The results indicate that performance remains largely stable across different $\lambda$ values, suggesting Mamba’s robustness to this hyper-parameter. Based on these findings, we set $\lambda = 0.5$ for all reported experiments.

\noindent\textbf{SSM state size.}
In Fig.~\ref{fig:state}, we assess the impact of varying SSM state size on the performance of our methods. We conducted experiments on ImageNet-1K and CIFAR-100, training the MambaCL with state sizes of 16, 32, 64, 128, and 256. The results demonstrate a consistent improvement in performance as the state size increases. To optimally balance performance with computational efficiency, we selected a state size of 128 for our experiments.

\noindent\textbf{Different architectures.} Table~\ref{tab:ab} presents an ablation study comparing Mamba-1, MambaFormer~\citep{park2024can}, and Mamba-2 within our MambaCL framework. MambaFormer is a hybrid model that integrates vanilla attention into Mamba and replaces Transformer positional encoding with a Mamba block. The results show MambaFormer performs comparably to Transformers, while Mamba-2 achieves the best performance on the CIFAR-100 dataset. \rev{Results with more attention-free architectures (\eg, RWKV and Longhorn) are provided in the appendix.}

\noindent\rev{\textbf{Mamba components.} Table~\ref{tab:component} further ablates the components of Mamba (\ie, Mamba-2). Both gating and short convolution improve performance; the short convolution accelerates convergence and stabilizes training, consistent with the shift-SSM interpretation in H3~\citep{fu2022hungry}, enabling state carry-over and comparison across nearby tokens. Removing the S6 (selective SSM) block entirely causes training failure with accuracy at the random-guess level ($5.0\%$, \ie, $1/20$), which persists even with increased depth ($4\rightarrow6$ layers) and width ($512\rightarrow768$): the short convolution (kernel size 4) is a purely local operation, and without the SSM the model cannot capture the long-range dependencies required by MCL. The selective SSM is thus essential, while convolution and gating provide complementary gains.}

\noindent\textbf{Mamba+MoE.}\label{sec:experiment_moe}
In Table~\ref{tab:ab}, we present experiments in which Mamba is augmented with mixture-of-experts (MoE), incorporating twelve 2-layer MLP expert networks with a dense-MoE router following each Mamba Block, resulting in improved performance. Additionally, we include performance metrics for vanilla and linear transformers as references.

\noindent\textbf{Computational cost.}
\rev{Measured with the official implementations under streaming inference, Mamba is more efficient than the vanilla Transformer in both parameters ($4.5$M vs. $8.4$M sequence-model parameters) and memory: its total inference memory stays \emph{constant} at $26.3$MB regardless of the stream length, whereas the Transformer's grows with the stream ($44.4$MB $\rightarrow$ $56.9$MB from 20-task 5-shot to 100-task 5-shot) as its key-value cache accumulates ($3.1$MB $\rightarrow$ $15.6$MB). Details are provided in the appendix.}

\section{Conclusion}
In this paper, we formulate SSMs and Mamba as sequence prediction-based continual learners, meta-trained on CL episodes with a selectivity regularization. Extensive experiments show that Mamba outperforms other attention-free methods and matches or surpasses Transformers using fewer parameters and \rev{constant memory}. It also exhibits strong reliability, generalization, and robustness under global structures, domain shifts, and long sequences.

\noindent\textbf{Limitations and future work.} This study can be extended to larger-scale datasets and more CL settings. Future work also includes exploring other types efficient models, such as efficient Transformers with compressed key-value caches and CL for foundation models tuning. Beyond the current MCL framework, we aim to expand its applicability to a broader range of scenarios and models.

{
    \small
    \bibliographystyle{ieeenat_fullname}
    \bibliography{main}
}

\appendix

\section{Datasets}
\label{sec:dataset_appendix}
\subsection{General Image Classification Tasks}
\textbf{CIFAR-100}~\citep{krizhevsky2009learning} consists of 60,000 images across 100 classes, each with 600 images.
\rev{We randomly select 40 classes for meta-testing and use the remaining 60 for meta-training.}

\noindent \textbf{ImageNet-1K}~\citep{russakovsky2015imagenet} comprises over one million labeled images across 1,000 categories. \rev{We randomly select 400 classes for meta-testing and use the remaining 600 for meta-training.}

\noindent\textbf{ImageNet-R}~\rev{\citep{hendrycks2021many}} extends 200 ImageNet classes with 30,000 renditions (\eg, art, sketches, and cartoons) tailored for robustness research. \rev{We randomly select 80 classes for meta-testing and use the remaining 120 for meta-training.}

\noindent\textbf{Celeb}~\citep{guo2016ms} is a large-scale facial image collection with approximately 10 million images of 100,000 celebrities. We randomly select 1,000 classes for meta-testing and use the remaining classes for meta-training.

\noindent \textbf{CASIA Chinese handwriting}~\citep{liu2011casia} encompasses 7,356 character classes with 3.9 million images. We randomly select 1,000 classes for meta-testing and use the remaining classes for meta-training.

\noindent\textbf{Omniglot}~\citep{lake2015human} is a collection of \rev{1,623} handwritten characters from 50 different alphabets, with 20 images per character.
The meta-training set comprises 963 classes, and the meta-testing set includes the remaining 660 classes.

\subsection{Fine-Grained Recognition Tasks}
\textbf{CUB-200-2011}~\citep{wah2011caltech} is a widely used fine-grained visual categorization dataset, comprising 11,788 images across 200 bird subcategories. We randomly select 80 classes for meta-testing and use the remaining \rev{120} for meta-training.

\noindent \textbf{Stanford Dogs}~\citep{dataset2011novel} comprises 20,580 images spanning 120 dog breeds from around the world. We randomly select 48 classes for meta-testing and use the remaining 72 for meta-training.

\noindent\textbf{Stanford Cars}~\citep{krause20133d} comprises 16,185 images across 196 car classes. We randomly select 80 classes for meta-testing and use the remaining \rev{116} for meta-training.

\noindent \textbf{FGVC-Aircraft}~\citep{maji2013fine} comprises 10,200 images across 102 aircraft model variants, each with 100 images. We randomly select 40 classes for meta-testing and use the remaining \rev{62} for meta-training.

\subsection{Large Domain Shift Tasks}
\textbf{DomainNet}~\citep{peng2019moment} is a benchmark for domain adaptation, encompassing common objects organized into 345 classes across six domains: clipart, real, sketch, infograph, painting, and quickdraw. 
We evaluate model adaptability to out-of-domain data by using one domain for meta-testing and the remaining domains for meta-training.

\subsection{Regression Tasks}
\textbf{Sine Wave Reconstruction (Sine).}
The sine wave $\omega(\tau) = A \sin(2\pi \nu \tau + \psi)$ is defined by its amplitude $A$, frequency $\nu$, and phase $\psi$.
We denote the target values $y$ as evaluations of the sine wave at 50 predefined points: $y = [\omega(\tau_1), \dots, \omega(\tau_{50})]$.
In each task, the frequency and phase remain constant, but the amplitude is allowed to vary.
To corrupt $y$ into $x$, we introduce a phase shift and Gaussian noise, where the phase shift is randomly selected for each task.
The mean squared error between $y$ and the model's prediction $\hat y$ is reported as the evaluation criterion.

\noindent\textbf{Image Rotation Prediction (Rotation).}
The model is provided with an image rotated by an angle $\theta \in [0, 2\pi)$, and its task is to predict the rotation angle $\hat{\theta}$. We use $1 - \cos(\hat{\theta} - \theta)$ as the evaluation metric, where a perfect prediction would result in a score of 0, while random guessing would yield an average score of 1.0. The CASIA dataset is employed, with each class treated as an individual task, maintaining the same meta-split configuration.

\noindent\textbf{Image Completion (Completion).}
In this task, the model fills in the missing part of an image given the visible part. Using the CASIA dataset, we take the top half of the image as the input $x$ and the bottom half as the target $y$. We report the mean squared error between $y$ and the model's prediction $\hat{y}$ as the evaluation criterion.

\begin{table}[t]
\caption{\rev{Model configurations of all compared methods; ``-'' denotes not applicable.}}
\label{tab:config}
\begin{center}
\small
\resizebox{0.68\linewidth}{!}
{
\begin{tabular}{lcccc}
\toprule

 &Mamba & Transformer & Linear TF & Performer \\
\midrule
Batch size       &\multicolumn{4}{c}{16}    \\
Max Train Step&\multicolumn{4}{c}{50000}    \\
Optimizer&\multicolumn{4}{c}{Adam}\\
Learning Rate &\multicolumn{4}{c}{$1\times10^{-4}$}    \\
Learning Rate Decay  &\multicolumn{4}{c}{Step}\\
Learning Rate Decay Step  &\multicolumn{4}{c}{10000}\\
Learning Rate Decay Rate  &\multicolumn{4}{c}{0.5}\\
Regularization $\lambda$  &\multicolumn{4}{c}{0.5} \\
Hidden Dimension&\multicolumn{4}{c}{512}\\
Layer&\multicolumn{4}{c}{4}\\

State Size&{128}&-&-&-\\
\rev{Conv Kernel Size}&4&-&-&-\\
\rev{Expand Factor}&\rev{1}&-&-&-\\
\rev{Head Dimension}&\rev{128}&-&-&-\\
Attention&-&Softmax&Elu&Favor\\
\rev{Attention Heads}&-&\rev{8}&\rev{8}&\rev{8}\\
\rev{FFN Dimension}&-&\rev{1024}&\rev{1024}&\rev{1024}\\
\rev{Dropout}&-&\rev{0.1}&\rev{0.1}&\rev{0.1}\\

\bottomrule
\end{tabular}
}
\end{center}
\end{table}

\section{Implementation Details}
\label{sec:implementation_details}

\begin{wrapfigure}{r}{0.30\linewidth}
\vspace{-18pt}
    \centering
    \includegraphics[width=0.92\linewidth]{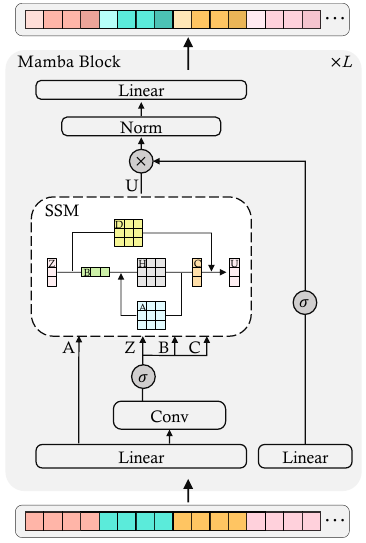}
    \caption{The Mamba block in MambaCL.}
    \label{fig:MambaBlock}
\end{wrapfigure}

We conduct our main experiments \rev{in PyTorch} on an NVIDIA A100 GPU, repeating each experiment five times \rev{with random seeds 0--4} and reporting the mean and standard deviation at convergence of meta-training.
We use a batch size of 16 and the Adam optimizer, with an initial learning rate of $1\times10^{-4}$ that decays by $0.5$ every $10,000$ steps. All models follow a consistent setup to ensure fair comparisons, with details provided in \rev{Table~\ref{tab:config}}.

\par
\rev{For training from scratch, we follow the settings of \citet{lee2023recasting}. For experiments with pre-trained representations, we employ CLIP ViT-B/16~\citep{radford2021learning} as the image encoder, keeping its parameters frozen and adding a trainable linear projector.} \rev{Each image is encoded as a single token using the CLS-token representation of the encoder.}

\par
\rev{Fig.~\ref{fig:MambaBlock} illustrates the Mamba block used in our work: it applies a 1-D convolution on the input tokens and projects the representations to obtain the input-dependent SSM parameters \citep{dao2024transformers}; multiple Mamba blocks are stacked homogeneously.}

\begin{algorithm}[th]
\caption{\rev{Meta-testing (CL inference) with MambaCL}}
\label{alg:inference}
\begin{algorithmic}[1]
\REQUIRE Meta-trained Mamba $f_\theta$; CL data stream $\gD^\text{train}=(\rvx_1, y_1, \ldots, \rvx_T, y_T)$; test input $\rvx^\text{test}$
\ENSURE Prediction $\hat{y}^\text{test}$
\STATE Initialize hidden state $\rmH_0 \leftarrow \mathbf{0}$
\FOR{$t = 1$ \TO $T$}
\STATE $\rmH_{t-1}^{+} \leftarrow f_\theta(\rvx_t, \rmH_{t-1})$ ~~\COMMENT{take the input token}
\STATE $\rmH_{t} \leftarrow f_\theta(y_t, \rmH_{t-1}^{+})$ ~~\COMMENT{take the target token}
\ENDFOR
\STATE $\hat{y}^\text{test} \leftarrow f_\theta(\rvx^\text{test}, \rmH_T)$
\RETURN $\hat{y}^\text{test}$
\end{algorithmic}
\end{algorithm}

\begin{algorithm}[th]
\caption{\rev{Meta-training of MambaCL}}
\label{alg:training}
\begin{algorithmic}[1]
\REQUIRE Initial parameters $\theta$; episode distribution $P_{(\gX, \gY)}$; task loss $\ell$; regularization weight $\lambda$; learning rate $\beta$
\ENSURE Meta-learned parameters $\theta$
\WHILE{not converged}
\STATE Sample a CL episode $(\gD^\text{train}, \gD^\text{test}) \sim P_{(\gX, \gY)}$ with $\gD^\text{train}=(\rvx_1, y_1, \ldots, \rvx_T, y_T)$
\STATE $\rmH_0 \leftarrow \mathbf{0}$;~ $\ell_\text{slct} \leftarrow 0$
\FOR{$t = 1$ \TO $T$}
\STATE $(\rmH_{t-1}^{+}, \rmB_{2t-1}, \rmC_{2t-1}) \leftarrow f_\theta(\rvx_t, \rmH_{t-1})$
\STATE $(\rmH_{t}, \rmB_{2t}, \rmC_{2t}) \leftarrow f_\theta(y_t, \rmH_{t-1}^{+})$
\STATE $\rvq_{2t-1} \leftarrow [\rmC_{2t-1}\rmB_{j}^{\top}]_{j=1}^{2t-2}$;~ compute the association pattern $\rvp_{2t-1}$ from $y_t$ and $\{y_j\}_{j<t}$
\STATE $\ell_\text{slct} \leftarrow \ell_\text{slct} + \text{KL}(\rvp_{2t-1}, \rvq_{2t-1})$
\ENDFOR
\STATE $\gL \leftarrow \sum_{(\rvx^\text{test}, y^\text{test})\in \gD^\text{test}} \ell\big(f_\theta(\rvx^\text{test}, \rmH_T), y^\text{test}\big) + \lambda\, \ell_\text{slct}$
\STATE $\theta \leftarrow \theta - \beta\, \nabla_\theta \gL$
\ENDWHILE
\end{algorithmic}
\end{algorithm}

\begin{table}[th]
\caption{\rev{Classification accuracy (\%) on 20-task 5-shot MCL with models trained from scratch}, on general image classification tasks. \highred{Best} and \highblue{second best} results are in \highred{red} and \highblue{blue}, respectively.}
\label{tab:scratch}
\begin{center}
\small

    {
    \begin{tabular}{lcccc}
    \toprule
        Method &CIFAR-100 &Omniglot &Casia &Celeb\\
    \midrule
        OML        &$10.1^{\pm0.4}$ &$75.2^{\pm2.2}$ &$96.8^{\pm0.1}$ &$57.5^{\pm0.2}$ \\
        Transformer&\highblue{$17.2^{\pm0.8}$} &\highblue{$86.3^{\pm0.6}$} &\highred{$99.6^{\pm0.0}$} &\highred{$70.0^{\pm0.2}$}\\
        Linear TF  &$16.6^{\pm0.5}$ &$64.0^{\pm1.4}$ &$99.3^{\pm0.0}$ &${67.6}^{\pm0.3}$ \\
        Performer  &$17.1^{\pm0.3}$ &$62.9^{\pm4.6}$ &$99.3^{\pm0.0}$ &$66.3^{\pm0.2}$ \\
        Mamba      &\highred{$18.3^{\pm0.4}$} &\highred{$87.7^{\pm0.5}$} &\highblue{$99.5^{\pm0.1}$} &\highblue{$68.1^{\pm0.1}$} \\
    \bottomrule
    \end{tabular}
    }
    \end{center}
\end{table}

\begin{table}[h]
\caption{\rev{Classification accuracy (\%) of additional attention-free models and the sparse-attention NSA on 20-task 5-shot MCL with pre-trained representations; Mamba is listed for reference.}}
\label{tab:more_models}
\begin{center}
\small
{
\begin{tabular}{lcc}
\toprule
Method &CIFAR-100 &ImageNet-1K \\
\midrule
Mamba    &$67.1^{\pm0.4}$ &$93.6^{\pm0.2}$ \\
Longhorn &$68.9^{\pm0.3}$ &$93.7^{\pm0.1}$ \\
RWKV-4   &$52.8^{\pm0.9}$ &$89.1^{\pm0.5}$ \\
RWKV-7   &$69.1^{\pm0.2}$ &$94.2^{\pm0.1}$ \\
NSA      &$60.8^{\pm0.5}$ &$92.5^{\pm0.4}$ \\
\bottomrule
\end{tabular}
}
    \end{center}
    \vspace{-0.3cm}
\end{table}

\section{\rev{Pseudocode of MambaCL}}
\label{sec:algorithm_appendix}
\rev{Algorithm~\ref{alg:inference} summarizes the meta-testing (CL inference) procedure: the meta-learned Mamba $f_\theta$ performs CL by recurrently updating its constant-size hidden state over the data stream, without any parameter update, and multiple test inputs are evaluated from the same retained state $\rmH_T$. Algorithm~\ref{alg:training} summarizes the meta-training procedure, where the input-dependent SSM parameters $\rmB$ and $\rmC$ are additionally maintained (\emph{only during training}) to compute the selectivity regularization introduced in the ``Selectivity Regularization for Meta-Training'' section of the main paper; the regularization term is computed for $t>1$. For brevity, $f_\theta(\cdot, \rmH)$ denotes one step of the stacked Mamba blocks that consumes a single token and returns the updated hidden state (or, for the final test query, the prediction).}

\section{Additional Experiments and Analyses}

\subsection{\rev{Results of Training from Scratch}}
\label{sec:scratch_appendix}
\rev{Table~\ref{tab:scratch} compares the architectures on general image classification tasks when training from scratch, following the settings of \citet{lee2023recasting} (see Sec.~\ref{sec:implementation_details}). Mamba achieves the best results on CIFAR-100 and Omniglot, and remains competitive with the Transformer on Casia and Celeb. On CIFAR-100, all methods suffer from substantial meta-overfitting: the meta-testing accuracy stays below $19\%$ for every architecture, while the same models reach far higher scores on the remaining datasets. This may be attributed to the small number of meta-training classes (60), \ie, limited task diversity for meta-learning.}

\subsection{\rev{Results with More Attention-Free Models}}
\label{sec:more_models_appendix}

\rev{Our framework is architecture-agnostic within the family of sequence-prediction models. To validate that the effectiveness is not exclusive to Mamba, we apply the same MCL meta-training (including the selectivity regularization) to other recent attention-free architectures with constant-size hidden states, \ie, RWKV-4~\citep{peng2023rwkv}, RWKV-7~\citep{peng2025rwkv}, and Longhorn~\citep{liu2025longhorn}. For reference, we also include NSA~\citep{yuan2025native}, an efficient sparse-attention (\emph{not} attention-free) method, implemented with the widely adopted \texttt{flash-linear-attention} library. All models follow the same 20-task 5-shot setting with pre-trained representations as in Table~1 of the main paper. As shown in Table~\ref{tab:more_models}, RWKV-7 and Longhorn match or surpass Mamba, confirming that strong MCL performance is not exclusive to Mamba but shared by attention-free models with constant-size hidden states. The comparison within this family further echoes our analysis of selectivity. Longhorn derives its recurrent update from an online-learning objective and thus writes new information into its state in an input-dependent, associative manner; this naturally fits the sequence-prediction formulation of MCL and yields performance on par with Mamba. RWKV-7 replaces the input-independent decay of earlier RWKV versions with dynamic, input-dependent state transitions (a generalized delta rule); this selective updating, analogous to Mamba's, leads to the best results among the compared models. In contrast, RWKV-4 accumulates history through a learned but input-independent per-channel decay and therefore cannot decide what to retain or forget based on the incoming sample, falling clearly behind (\eg, $52.8\%$ on CIFAR-100). NSA restricts each query's attention to compressed and selected token blocks: although effective for long-context language modeling, such token-level sparsity can drop support samples needed by later queries, leaving NSA behind the selective attention-free models.}

\begin{figure}[t]
    \centering

     \begin{subfigure}[b]{0.245\textwidth}
        \centering
        \includegraphics[width=\textwidth]{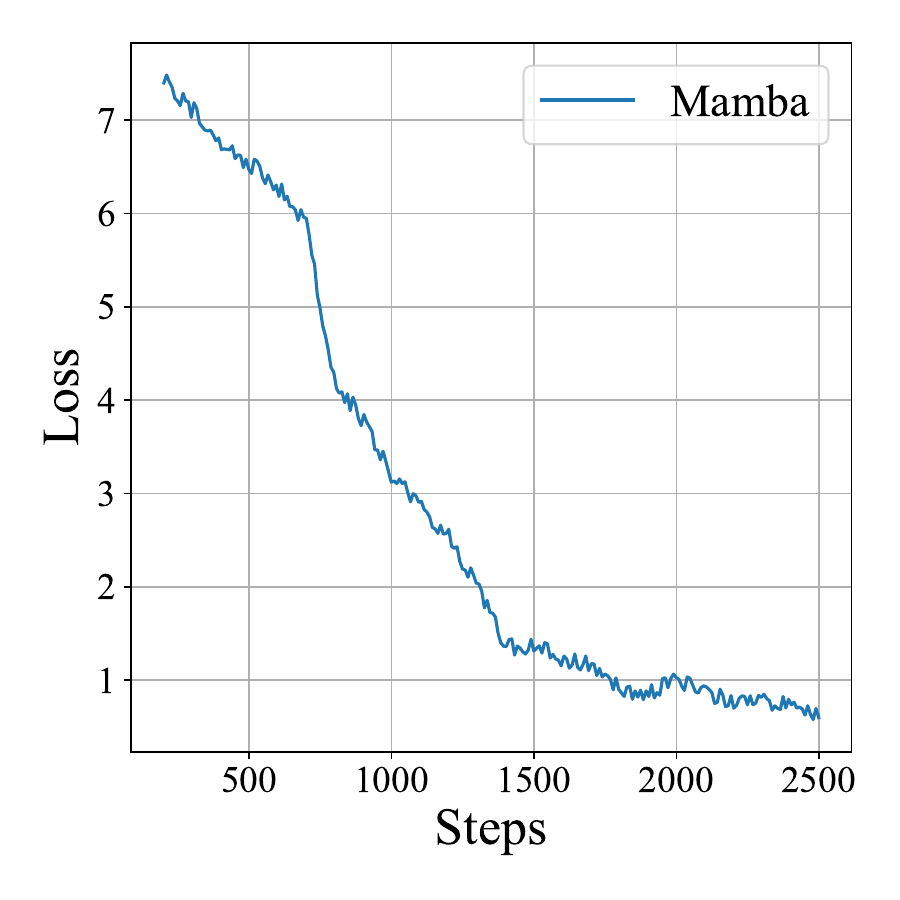}
        \caption{Mamba \textit{w/} $\ell_\text{slct}$}
    \end{subfigure}
    \hfill
    \begin{subfigure}[b]{0.245\textwidth}
        \centering
        \includegraphics[width=\textwidth]{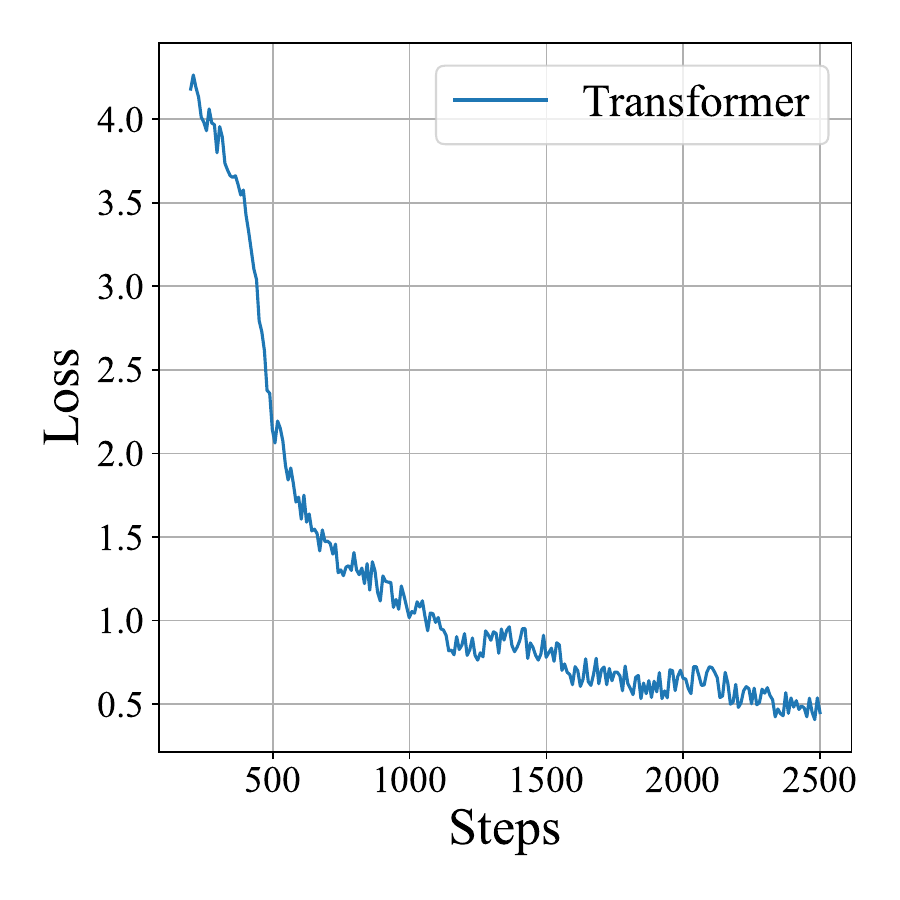} 
        \caption{Transformer \textit{w/} $\ell_\text{slct}$}
    \end{subfigure}
    \hfill
    \begin{subfigure}[b]{0.245\textwidth}
        \centering
        \includegraphics[width=\textwidth]{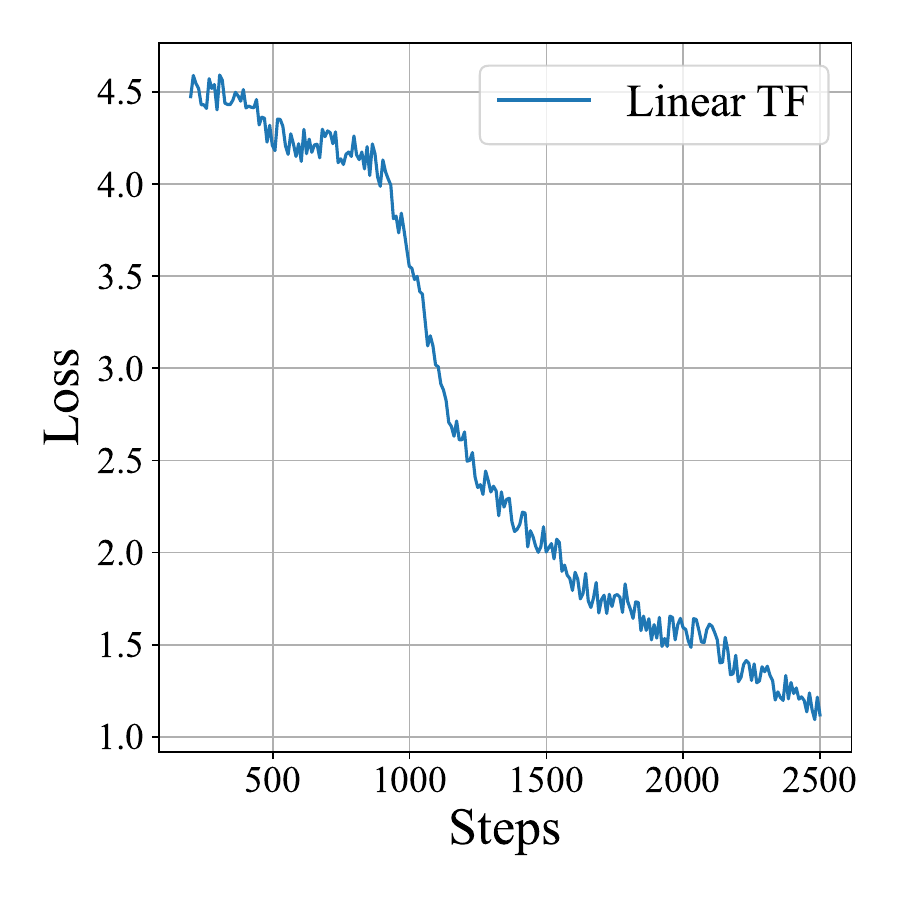} 
        \caption{Linear TF \textit{w/} $\ell_\text{slct}$}
    \end{subfigure}
    \hfill
    \begin{subfigure}[b]{0.245\textwidth}
        \centering
        \includegraphics[width=\textwidth]{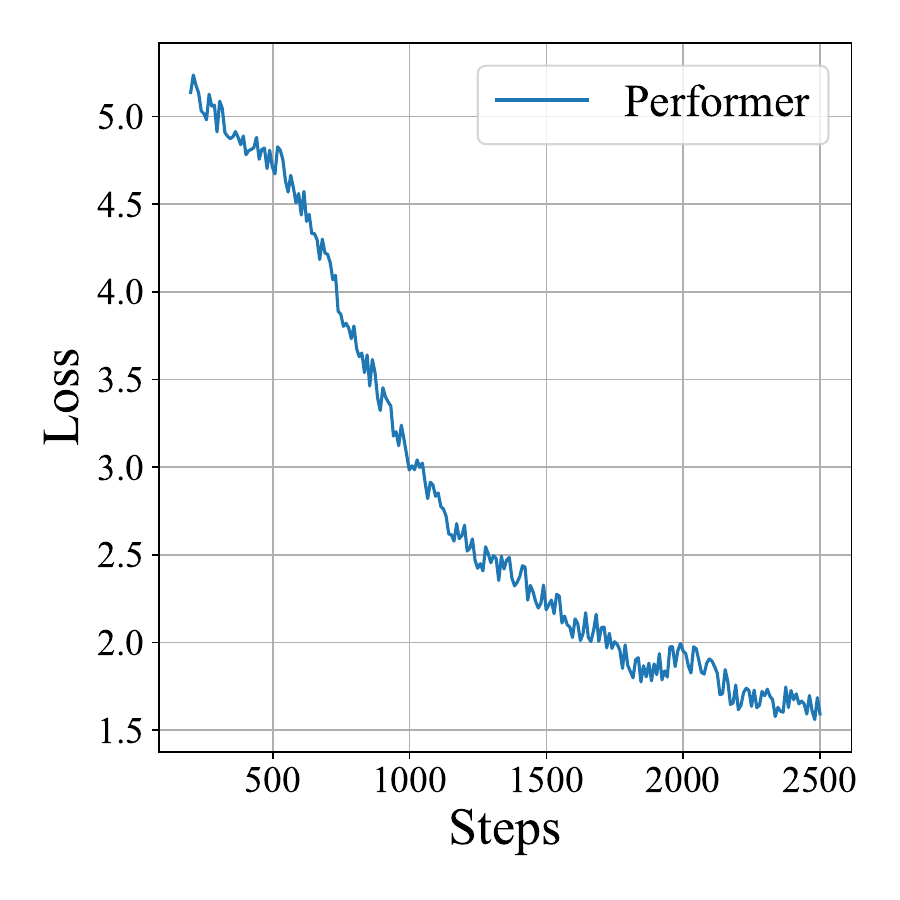} 
        \caption{Performer \textit{w/} $\ell_\text{slct}$}
    \end{subfigure}

    \begin{subfigure}[b]{0.245\textwidth}
        \centering
        \includegraphics[width=\textwidth]{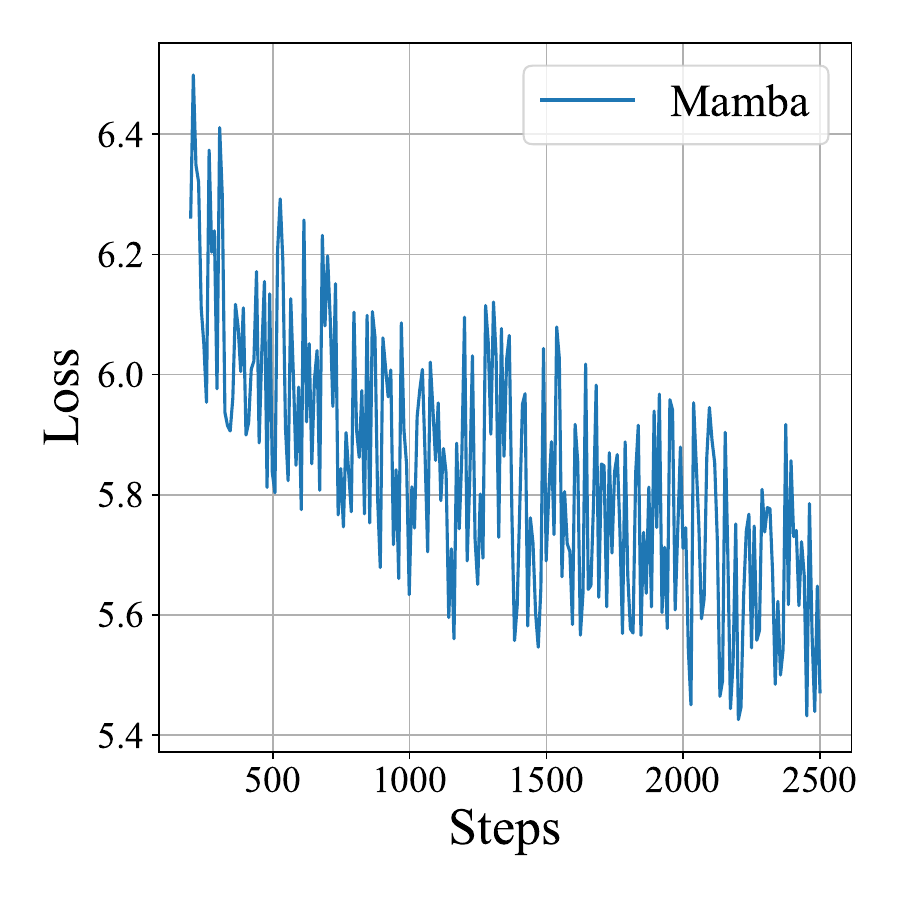} 
        \caption{Mamba \textit{w/o} $\ell_\text{slct}$}
    \end{subfigure}
    \hfill
    \begin{subfigure}[b]{0.245\textwidth}
        \centering
        \includegraphics[width=\textwidth]{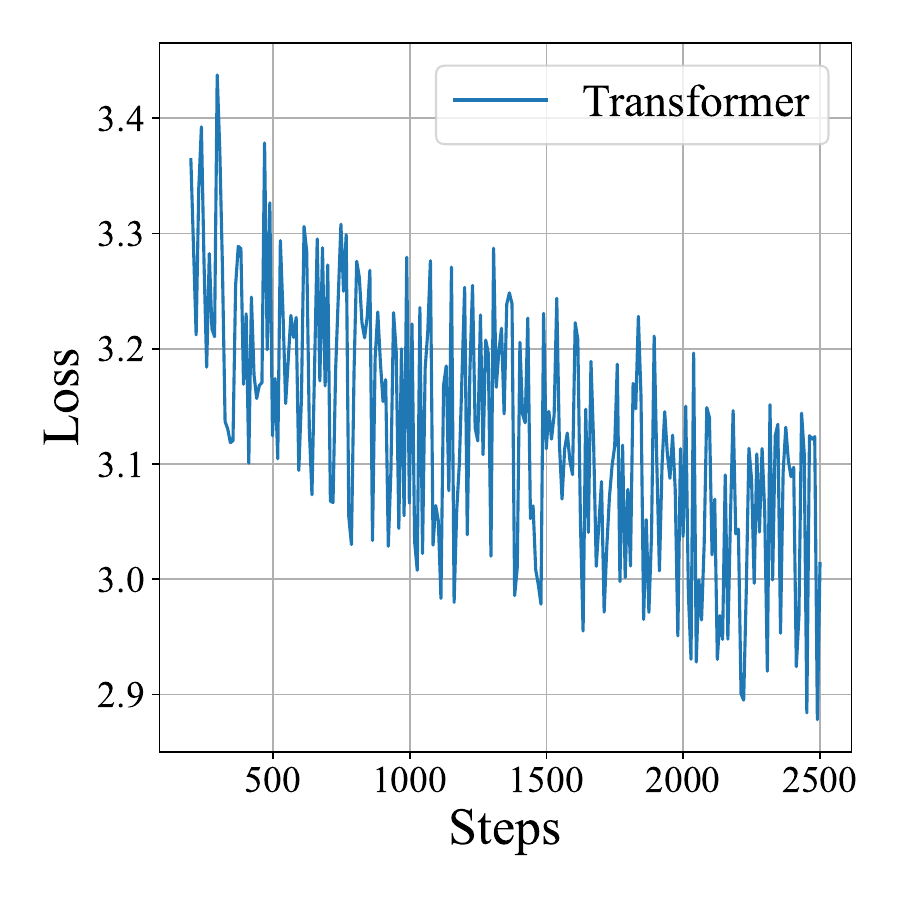} 
        \caption{Transformer \textit{w/o} $\ell_\text{slct}$}
    \end{subfigure}
    \hfill
    \begin{subfigure}[b]{0.245\textwidth}
        \centering
        \includegraphics[width=\textwidth]{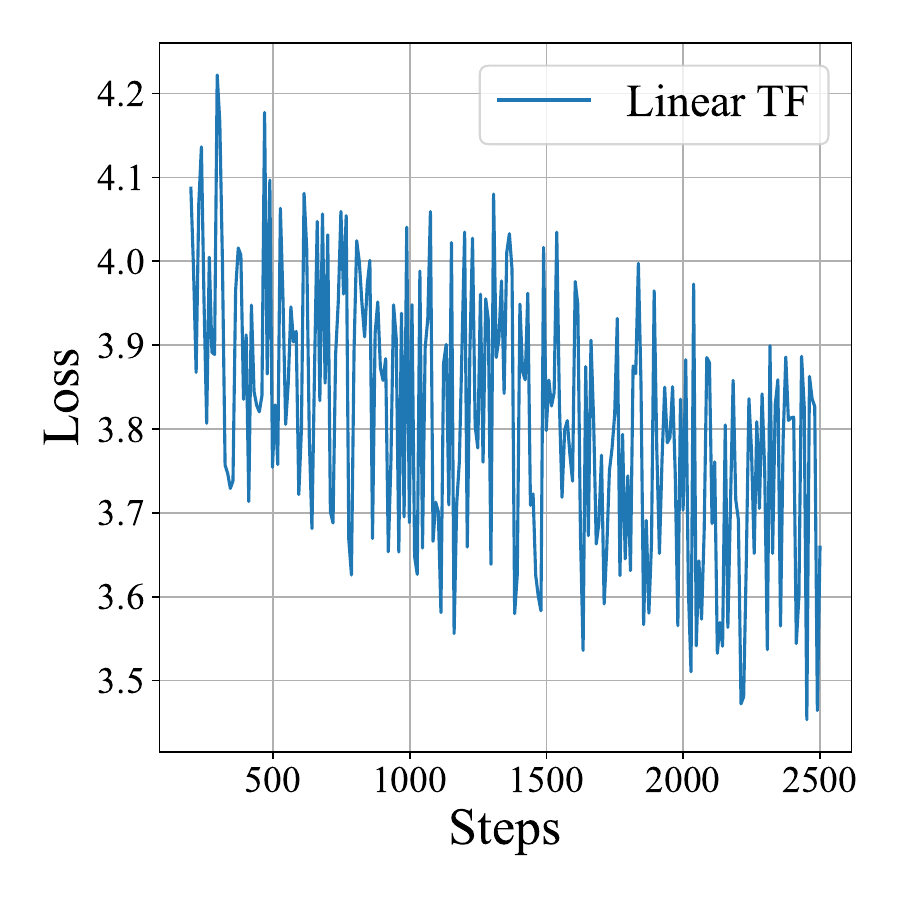} 
        \caption{Linear TF \textit{w/o} $\ell_\text{slct}$}
    \end{subfigure}
    \hfill
    \begin{subfigure}[b]{0.245\textwidth}
        \centering
        \includegraphics[width=\textwidth]{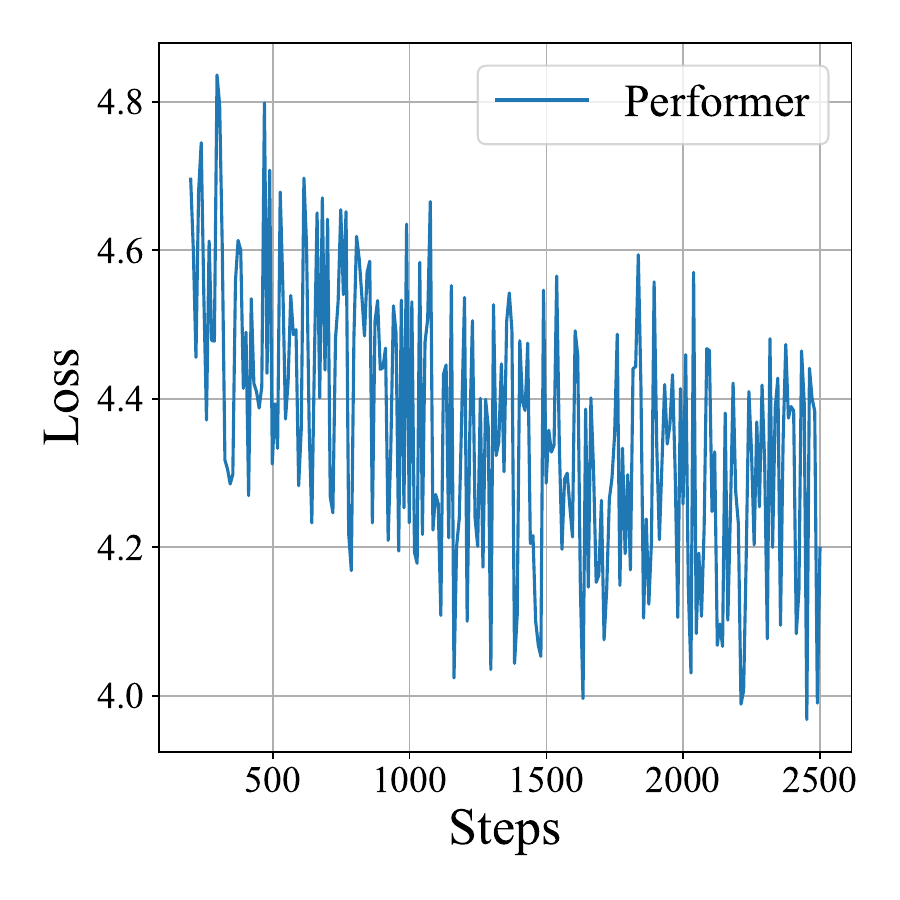} 
        \caption{Performer \textit{w/o} $\ell_\text{slct}$}
    \end{subfigure}

    \caption{\rev{Meta-training loss curves with (top, a--d) and without (bottom, e--h) the selectivity regularization ($\ell_\text{slct}$) for Mamba, Transformer, Linear Transformer, and Performer on 20-task 5-shot MCL with CIFAR-100. All models share the same input representations and training settings.}}
    \label{fig:loss_appendix}
\end{figure}

\begin{figure}[!th]
    \centering

     \begin{subfigure}[b]{0.245\textwidth}
        \centering
        \includegraphics[width=\textwidth]{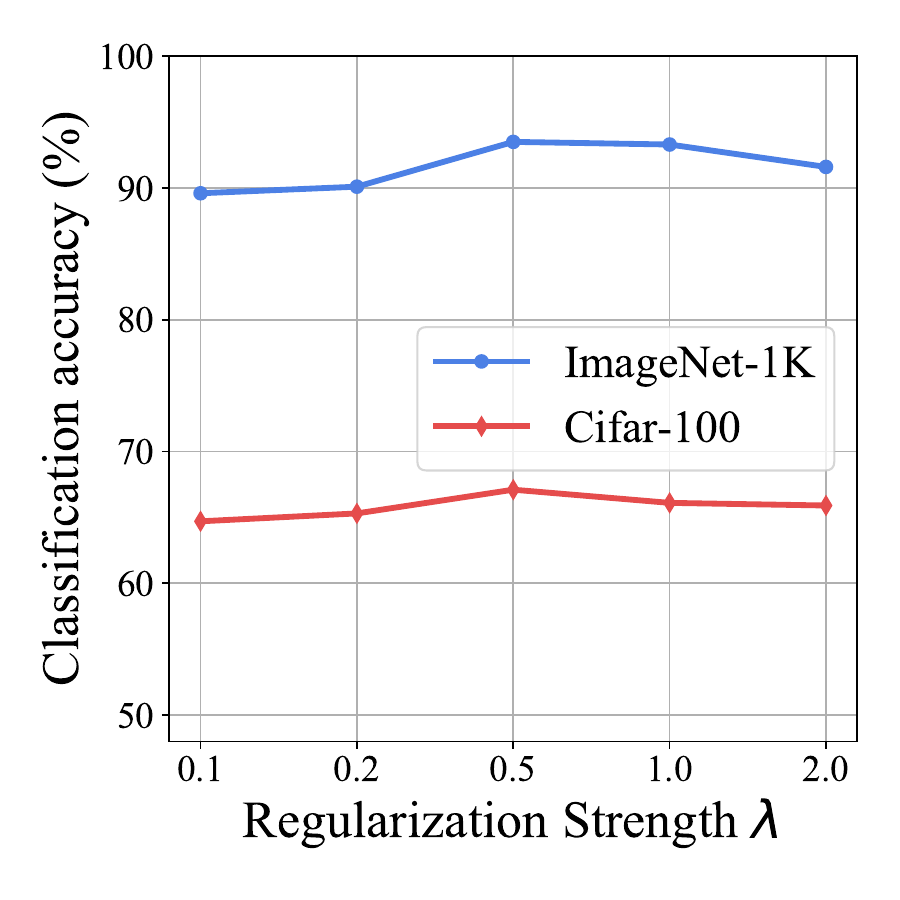} 
        \caption{Mamba}
    \end{subfigure}
    \hfill
    \begin{subfigure}[b]{0.245\textwidth}
        \centering
        \includegraphics[width=\textwidth]{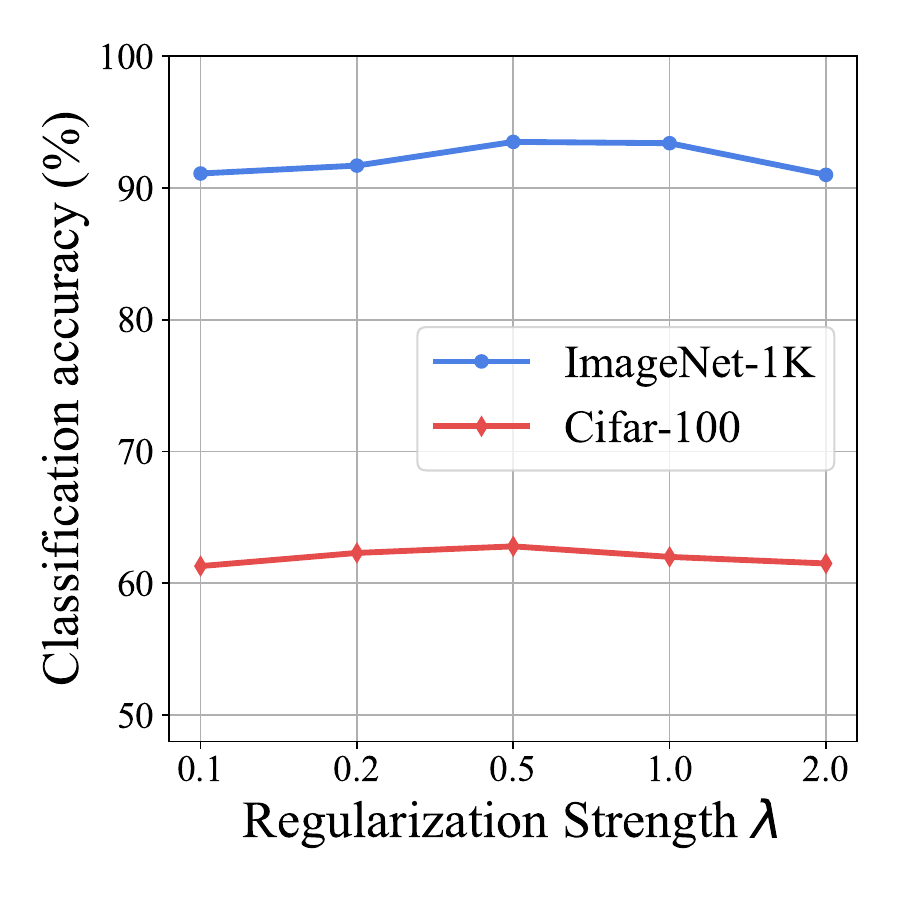} 
        \caption{Transformer}
    \end{subfigure}
    \hfill
    \begin{subfigure}[b]{0.245\textwidth}
        \centering
        \includegraphics[width=\textwidth]{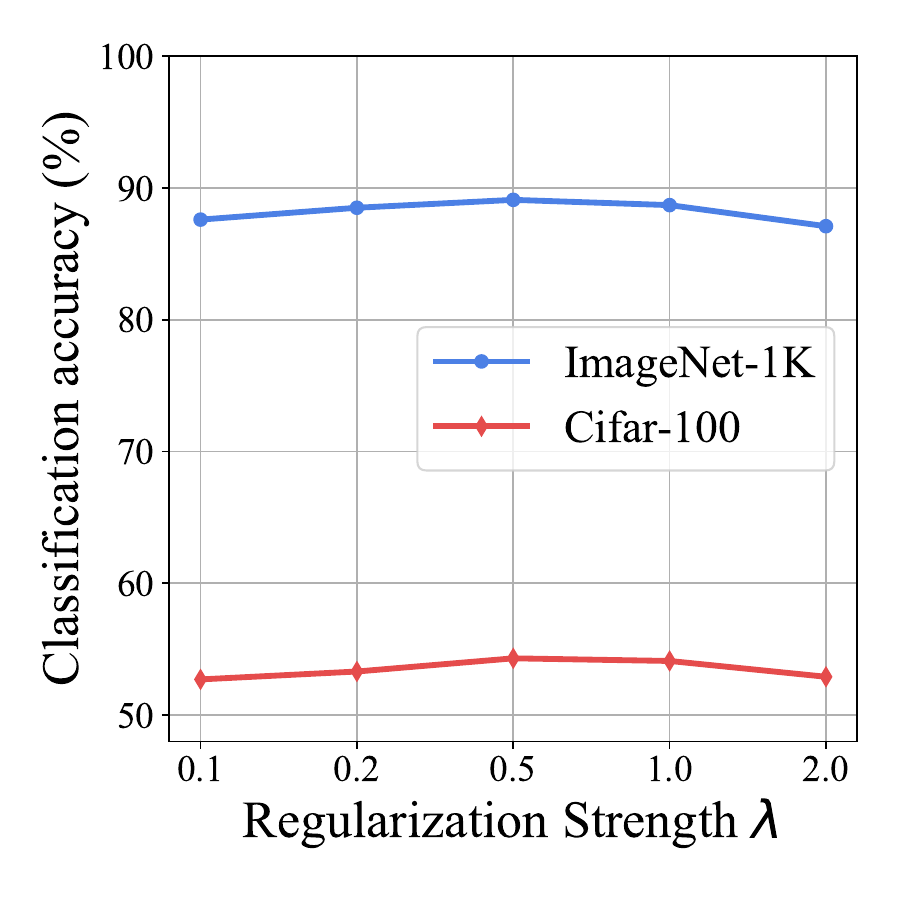} 
        \caption{Linear TF}
    \end{subfigure}
    \hfill
    \begin{subfigure}[b]{0.245\textwidth}
        \centering
        \includegraphics[width=\textwidth]{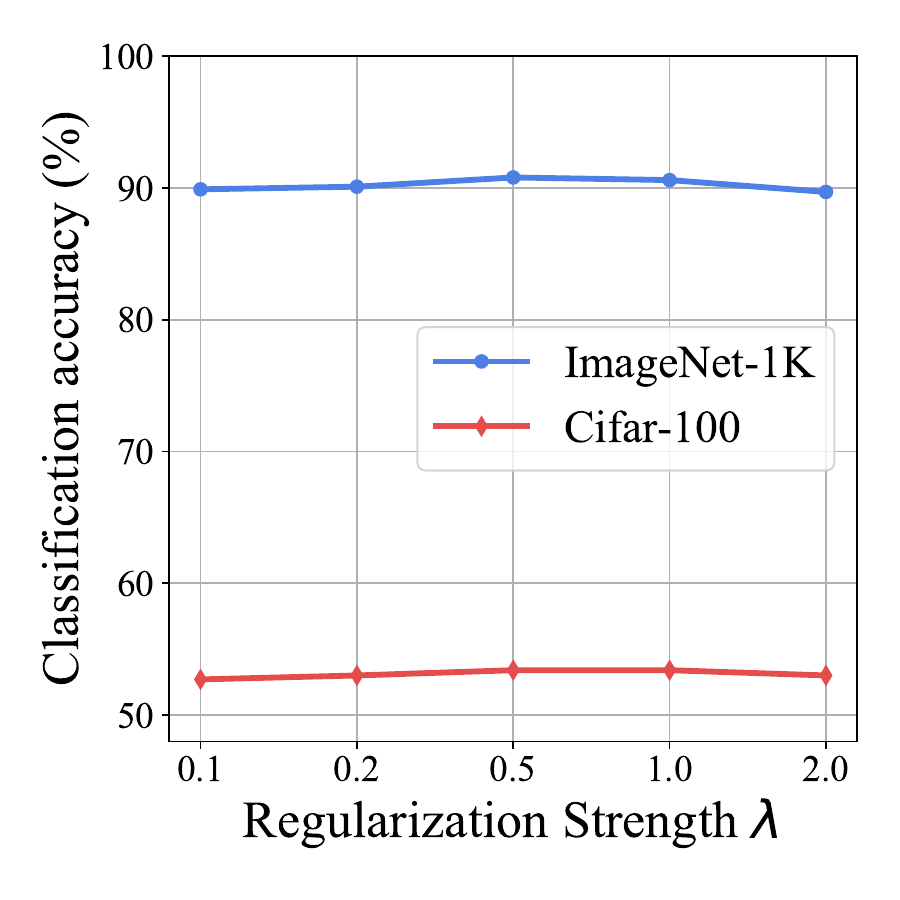} 
        \caption{Performer}
    \end{subfigure}
    \caption{\rev{Ablation on the regularization strength $\lambda \in \{0.1, 0.2, 0.5, 1.0, 2.0\}$ for (a) Mamba, (b) Transformer, (c) Linear Transformer, and (d) Performer, on CIFAR-100 and ImageNet-1K under 20-task 5-shot MCL.}}
    \label{fig:lambda_appendix}
\end{figure}

\begin{figure*}[t]
  \centering
  \begin{minipage}{0.28\textwidth}
      \centering
      \includegraphics[width=\linewidth]{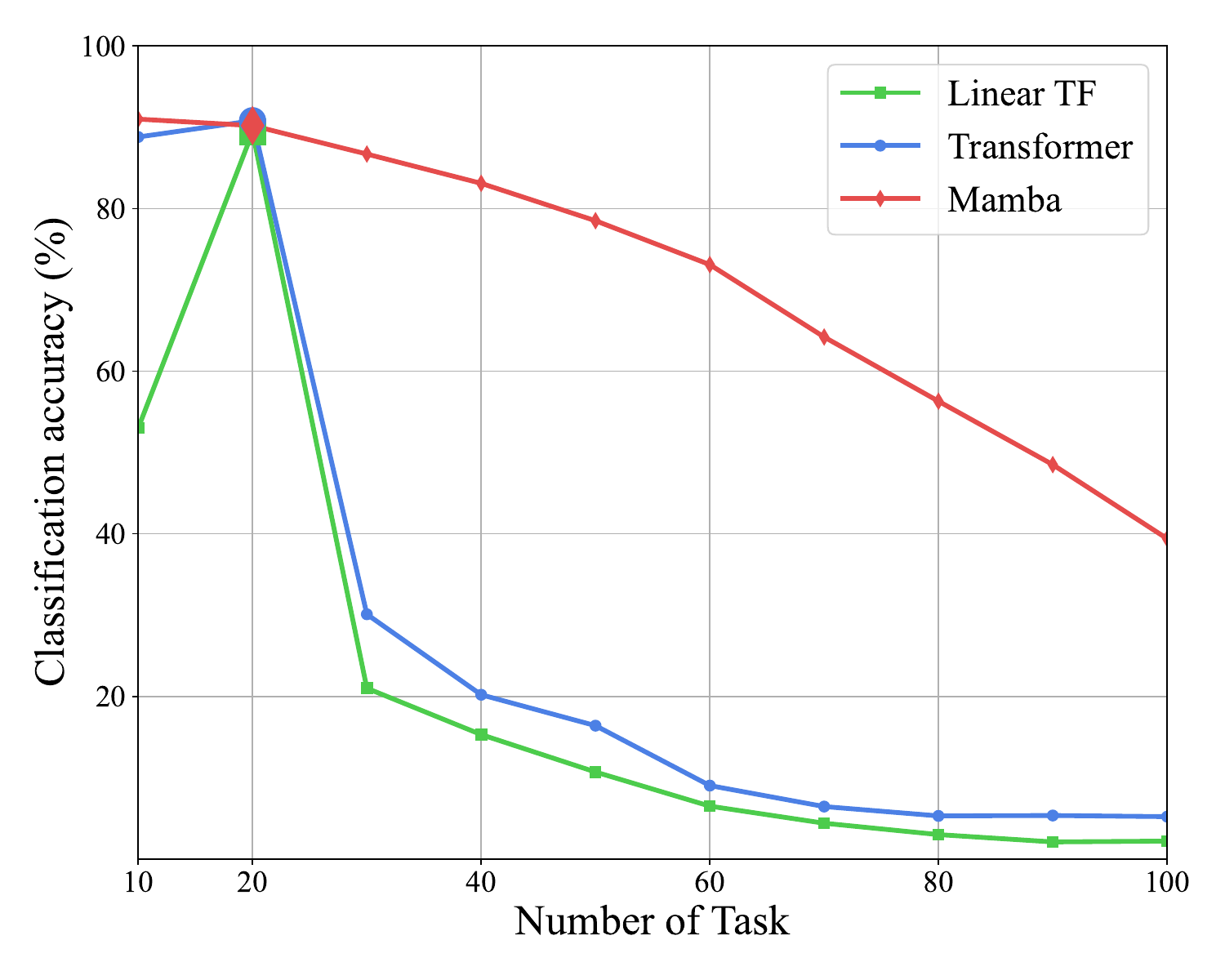}
      \subcaption{\small Task}
  \end{minipage}
  \begin{minipage}{0.28\textwidth}
      \centering
      \includegraphics[width=\linewidth]{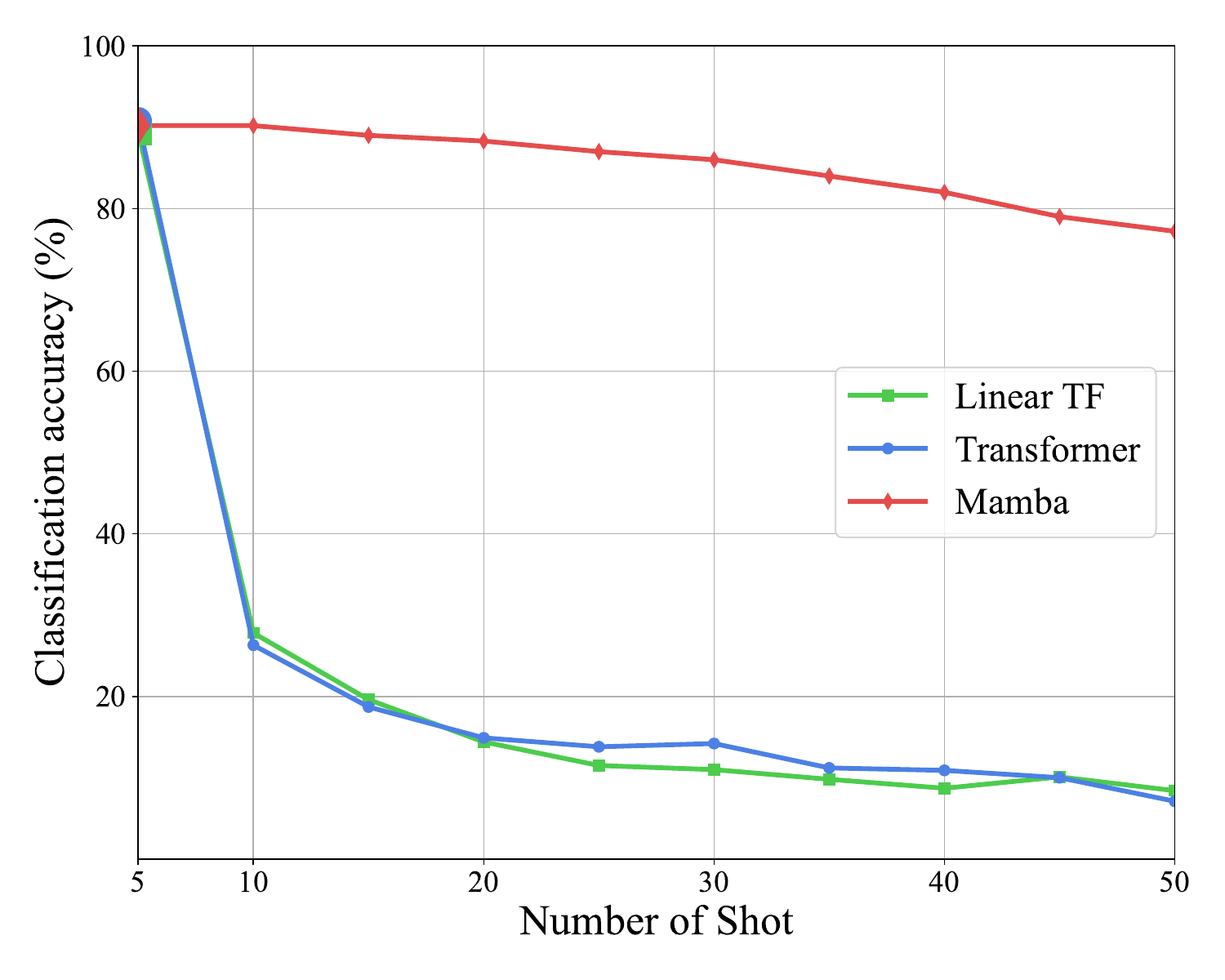}
      \subcaption{\small Shot}
  \end{minipage}
  \begin{minipage}{0.28\textwidth}
      \centering
      \includegraphics[width=\linewidth]{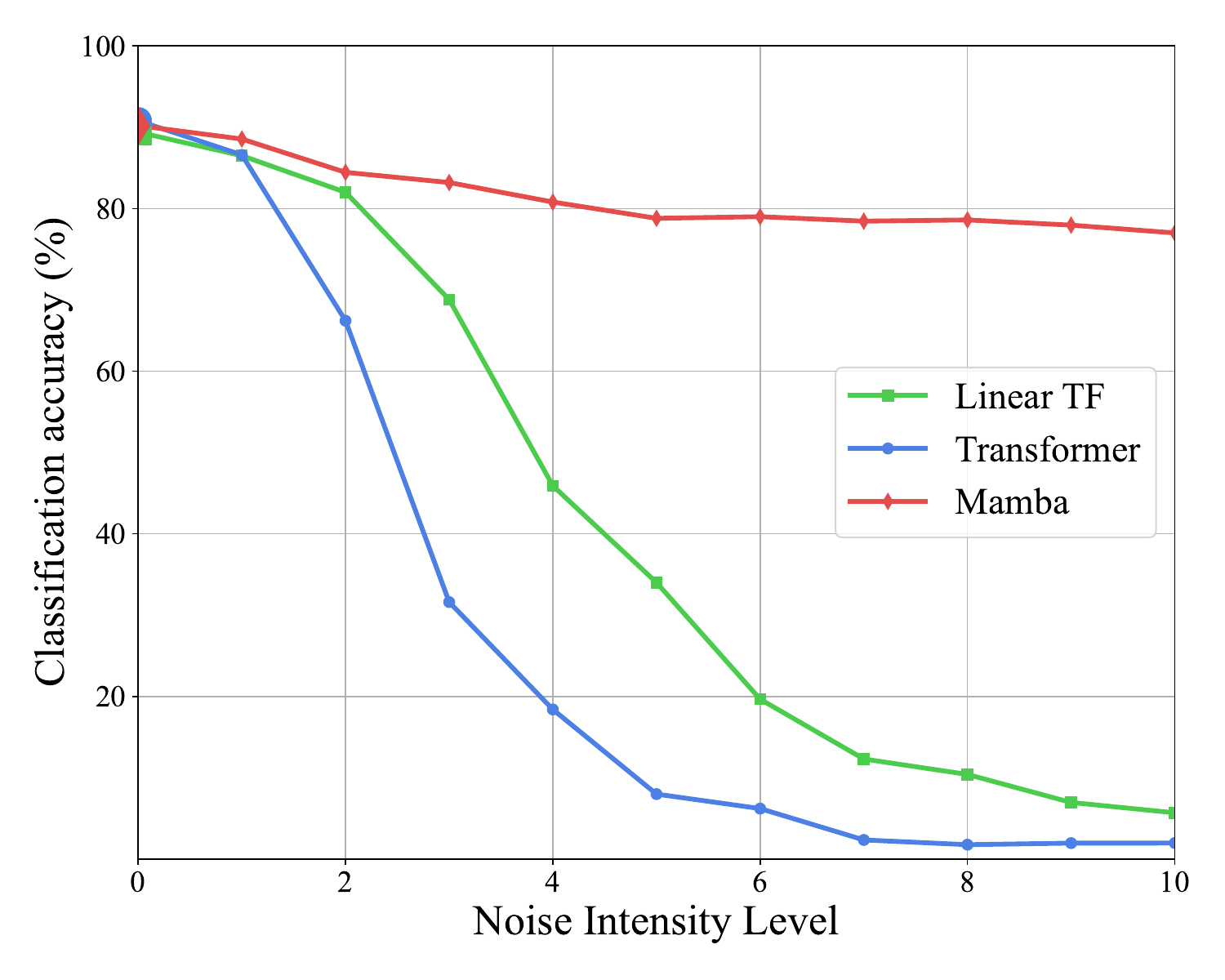}
      \subcaption{\small Noise}
  \end{minipage}
  \caption{\rev{Generalization analysis on ImageNet-1K with regularization strength $\lambda=0.1$ (the $\lambda=0.5$ counterpart is Fig.~3 of the main paper), with meta-training on 20 tasks (5-shot), meta-testing on: (a) varying tasks (5-shot), (b) varying shots (20-task), and (c) varying input noise levels (20-task 5-shot).}}
  \label{fig:generalization_appendix}
\end{figure*}

\begin{figure}[t]
    \centering

     \begin{subfigure}[b]{0.245\textwidth}
        \centering
        \includegraphics[width=\textwidth]{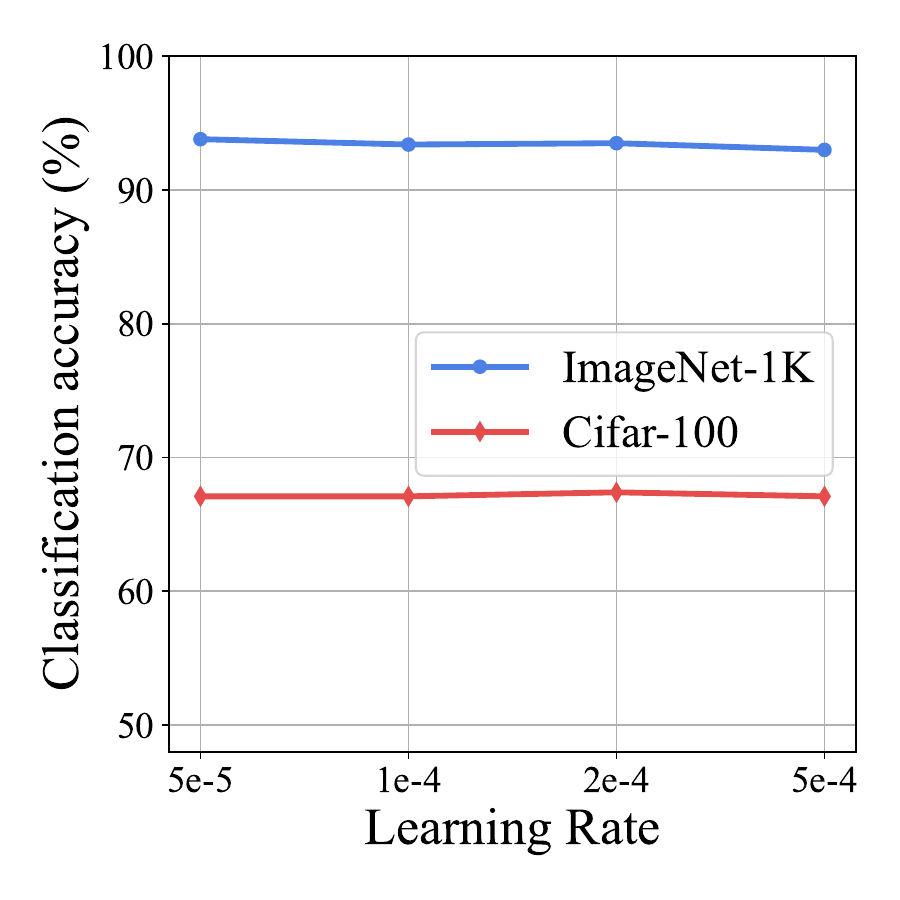} 
        \caption{Mamba}
    \end{subfigure}
    \hfill
    \begin{subfigure}[b]{0.245\textwidth}
        \centering
        \includegraphics[width=\textwidth]{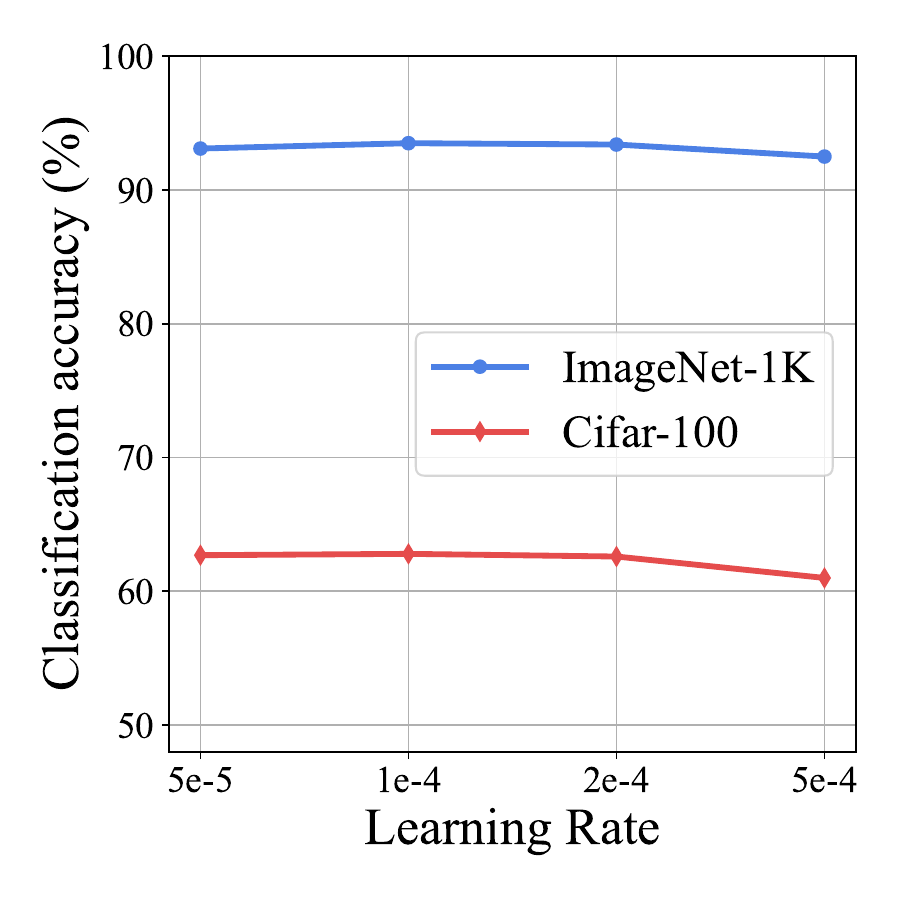} 
        \caption{Transformer}
    \end{subfigure}
    \hfill
    \begin{subfigure}[b]{0.245\textwidth}
        \centering
        \includegraphics[width=\textwidth]{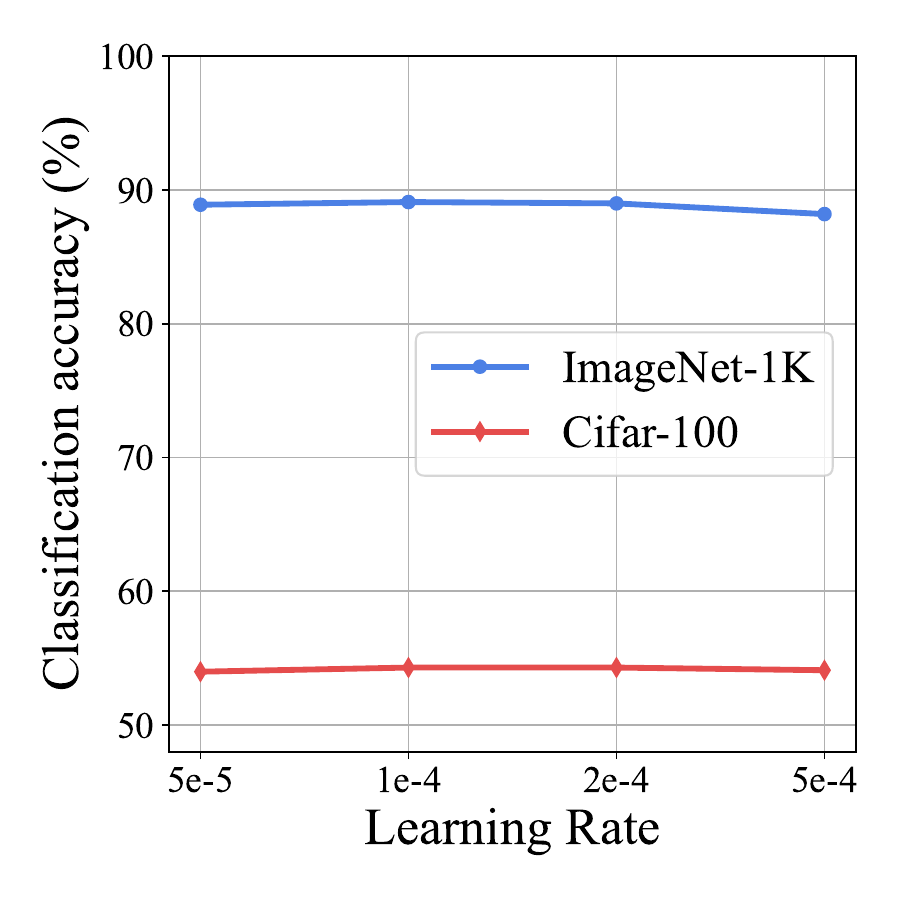} 
        \caption{Linear TF}
    \end{subfigure}
    \hfill
    \begin{subfigure}[b]{0.245\textwidth}
        \centering
        \includegraphics[width=\textwidth]{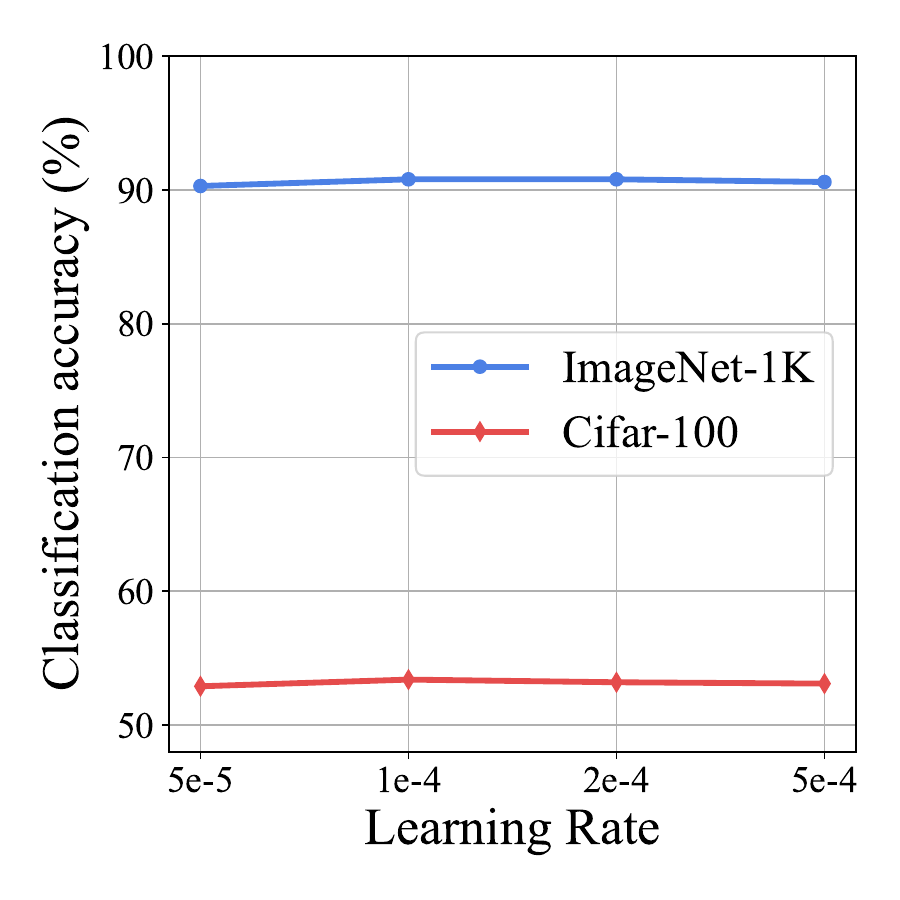} 
        \caption{Performer}
    \end{subfigure}
    \caption{\rev{Ablation on initial learning rates $\{5\times10^{-5}, 1\times10^{-4}, 2\times10^{-4}, 5\times10^{-4}\}$ for (a) Mamba, (b) Transformer, (c) Linear Transformer, and (d) Performer, on CIFAR-100 and ImageNet-1K under 20-task 5-shot MCL.}}
    \label{fig:lr_appendix}
\end{figure}

\begin{table}[h]
\caption{\rev{Computational cost of streaming inference on one 20-task 5-shot ($L{=}201$) or 100-task 5-shot ($L{=}1001$) MCL episode: sequence-model parameters, peak GPU memory, and the size of the retained state (Mamba) or key-value cache (Transformer). Mamba's memory footprint is constant in the stream length, whereas the Transformer's grows with it.}}
\label{tab:cost}
\begin{center}
\small
\resizebox{0.7\linewidth}{!}
{
\begin{tabular}{lccccc}
\toprule
\multirow{2}{*}{Method} &\multirow{2}{*}{Params.${\downarrow}$} &\multicolumn{2}{c}{Peak Memory${\downarrow}$} &\multicolumn{2}{c}{State/Cache${\downarrow}$} \\
 & &20t5s &100t5s &20t5s &100t5s \\
\midrule
Transformer &$8.4$M &$44.4$MB &$56.9$MB &$3.1$MB &$15.6$MB \\
Mamba &$4.5$M &$26.3$MB &$26.3$MB &$1.1$MB &$1.1$MB \\
Mamba (expand$=2$) &$7.7$M &$39.9$MB &$39.9$MB &$2.1$MB &$2.1$MB \\
\bottomrule
\end{tabular}
}
    \end{center}
\end{table}

\subsection{Effects of Selectivity Regularization on Meta-Training Loss Curves}
\label{sec:loss_appendix}
\rev{Fig.~\ref{fig:loss_appendix} shows the initial 2,500 meta-training steps of the four sequence models with and without the selectivity regularization ($\ell_\text{slct}$) on CIFAR-100. Without the regularization, the training losses stay 3--5 times higher than those of the regularized counterparts and, beyond 2,500 steps, oscillate without further decrease; with the regularization, all models converge stably. These results indicate that the selectivity regularization is crucial for stabilizing meta-training in the challenging MCL setting.}

\subsection{Ablation Studies on Regularization Strength}

\rev{Complementing the ablation on Mamba in Fig.~4a of the main paper, Fig.~\ref{fig:lambda_appendix} varies the regularization strength $\lambda \in \{0.1, 0.2, 0.5, 1.0, 2.0\}$ for all four sequence models on CIFAR-100 and ImageNet-1K. The performance of every model remains stable across this range, indicating insensitivity to the choice of $\lambda$; we therefore set $\lambda=0.5$ for all models by default.}

\subsection{Additional Generalization Analyses}

\rev{In the ``Generalization Analyses'' section and Fig.~3 of the main paper, we evaluate models meta-trained on 20-task 5-shot episodes under meta-testing episodes that differ from the meta-training configuration, \ie, with varying numbers of tasks or shots, or with noise-contaminated inputs. Mamba generalizes better to these unseen scenarios, whereas the Transformer exhibits stronger meta-overfitting. To verify that this gap is not an artifact of the regularization strength, Fig.~\ref{fig:generalization_appendix} repeats the generalization experiments with a small strength of $\lambda=0.1$: the phenomena closely match those under $\lambda=0.5$ (Fig.~3 in the main paper) across all models.}

\subsection{Ablation Studies on Learning Rates}
\label{sec:lr_appendix}
\rev{Fig.~\ref{fig:lr_appendix} varies the initial learning rate in $\{5\times10^{-5}, 1\times10^{-4}, 2\times10^{-4}, 5\times10^{-4}\}$ for all four sequence models on CIFAR-100 and ImageNet-1K. Within this range, the learning rate has little effect on the final performance; we therefore set it to $1\times10^{-4}$, decayed by $0.5$ every $10,000$ steps, for all models.}

\subsection{\rev{Computational Cost and Memory Growth}}
\label{sec:cost_appendix}
\rev{Table~\ref{tab:cost} reports the computational cost of the sequence-model backbones (configured as in Table~\ref{tab:config}), measured with the official Mamba-2 implementation (\texttt{mamba-ssm}) and a standard PyTorch attention implementation, in fp32 with batch size 1 on an NVIDIA RTX 4090. Following online CL semantics, each model processes one episode in a streaming fashion: input tokens arrive one at a time from the host and predictions are moved back, so the GPU holds only the model weights, the retained state (Mamba's recurrent state or the Transformer's key-value cache), and per-step activations. We report the number of sequence-model parameters (excluding the shared frozen encoder and projector), the peak GPU memory for inferring one episode, and the size of the retained state or cache, on 20-task 5-shot ($L{=}201$) and 100-task 5-shot ($L{=}1001$) episodes, together with a width-matched Mamba variant (expand${=}2$, \ie, doubled inner width) whose parameter count is comparable to the Transformer's.}

\rev{Mamba's total memory footprint remains \emph{constant} regardless of the stream length, whereas the Transformer's grows linearly with the stream. This is not merely an implementation detail. Under the sequence-prediction formulation of MCL, the retained state is exactly the ``memory'' of the continual learner: the Transformer keeps the key-value pairs of \emph{all} seen tokens, while Mamba compresses the stream into a constant-size hidden state. The reported memory thus reflects whether a learner satisfies CL's requirement of learning from a stream without storing all seen samples, and the gap between the two models widens as episodes become longer.}

\section{Visualization of Attention and Selectivity Patterns}
\label{sec:vis_appendix}
\rev{Meta-learned sequence models perform CL by processing the samples of a meta-testing episode in sequence: the prediction for a test input $\rvx^\text{test}$ is produced through the attention over, or the implicit association with, the samples seen in the stream. To analyze this behavior, we visualize the attention weights of the Transformer and the associative weights of Mamba (as discussed in the ``Selectivity Regularization for Meta-Training'' section of the main paper), which empirically show how the models make predictions. For the standard benchmarking case, Fig.~\ref{fig:20t5s_appendix} shows that both models effectively associate the test query with the seen samples of its class, consistent with the results in Table~1 of the main paper. We further use these visualizations to analyze how the models behave in the generalization studies (the ``Generalization Analyses'' section of the main paper), \ie, on meta-testing episodes that differ from those in meta-training.}

\subsection{Visualization Analysis of Generalization to Different Stream Lengths}

\rev{Figs.~3a and 3b of the main paper evaluate generalization by meta-testing on episodes longer than those seen during meta-training: models meta-trained on 20-task 5-shot episodes are meta-tested with more tasks or more shots. Transformers generally converge more easily than Mamba during meta-training owing to their strong fitting ability, but this advantage can also translate into meta-overfitting.}

\rev{To examine how the models handle such sequences, we visualize the final-layer attention weights of the Transformer and the corresponding selectivity scores (associative indicators) of Mamba between each test shot (query) and a single MCL training episode (prompt). Since Mamba produces no explicit attention weights, we rely on its connection to attention described in the ``Selectivity Regularization for Meta-Training'' section of the main paper: we compute $\rmC_\text{test}\rmB^{\top}$ within the SSM as the counterpart of the Transformer's attention matrix $\rmQ_\text{test}\rmK^{\top}$, where $\rmC_\text{test} \in \sR^{1\times C}$ and $\rmQ_\text{test} \in \sR^{1\times C}$ are the rows corresponding to the test shot. Both models are meta-trained on the 20-task 5-shot setting with ImageNet-1K.}

\begin{figure}[t]
    \begin{center}
    \begin{subfigure}[b]{\textwidth}
            \includegraphics[width=\linewidth]{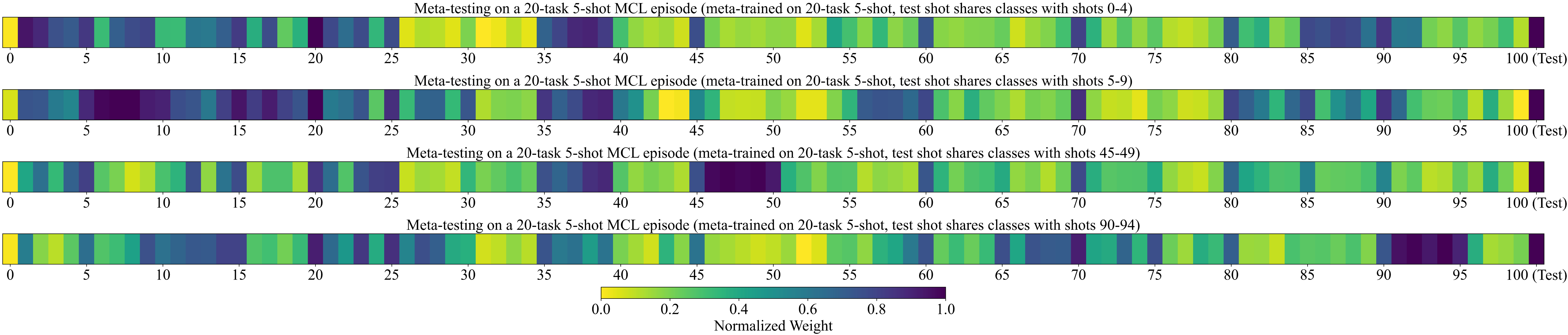}
    \caption{{Mamba}}
    \end{subfigure} \\

    \begin{subfigure}[b]{\textwidth}
            \includegraphics[width=\linewidth]{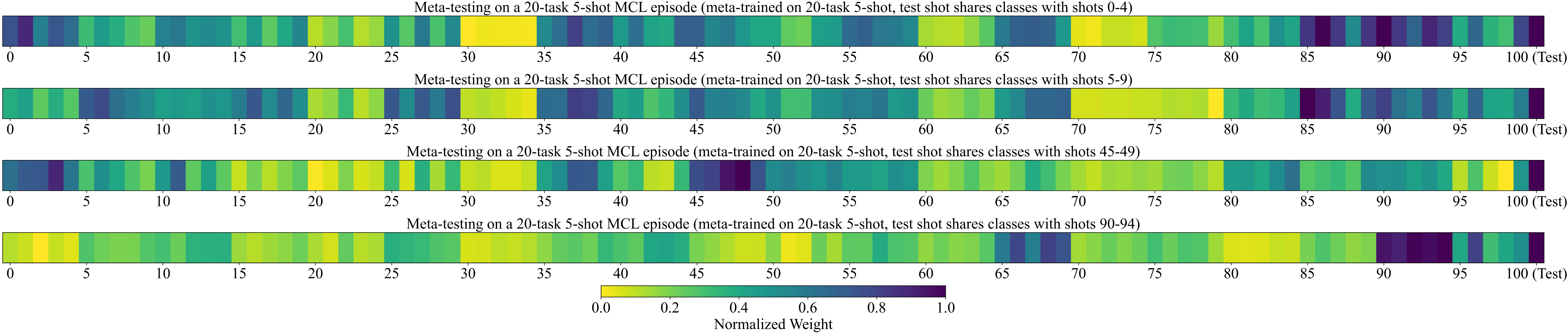}
    \caption{{Transformer}}
    \end{subfigure}
    \caption{\rev{Standard 20-task 5-shot meta-testing: final-layer associations between test shots (queries) and a single MCL training episode (prompt) for (a) Mamba and (b) Transformer, meta-trained and meta-tested on 20-task 5-shot MCL. The four visualizations share one training episode spanning the $0^{th}\!\!-\!99^{th}$ shots, and the queries correspond to the $0^{th}$, $1^{st}$, $9^{th}$, and $18^{th}$ tasks (the $0^{th}\!\!-\!4^{th}$, $5^{th}\!\!-\!9^{th}$, $45^{th}\!\!-\!49^{th}$, and $90^{th}\!\!-\!94^{th}$ training shots), respectively.}}
    \label{fig:20t5s_appendix}
    
    \end{center}
\end{figure}

\begin{figure}[th]
    \begin{center}
    \begin{subfigure}[b]{\textwidth}
            \includegraphics[width=\linewidth]{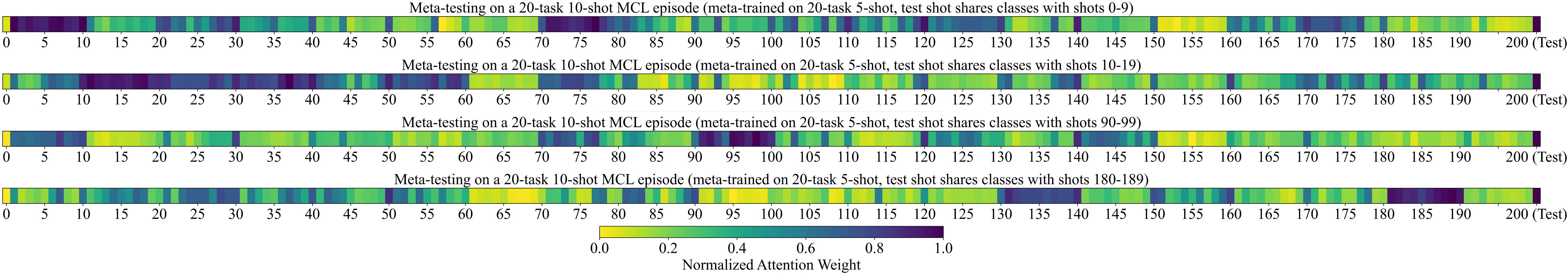}
    \caption{{Mamba}}
    \end{subfigure} \\

    \begin{subfigure}[b]{\textwidth}
            \includegraphics[width=\linewidth]{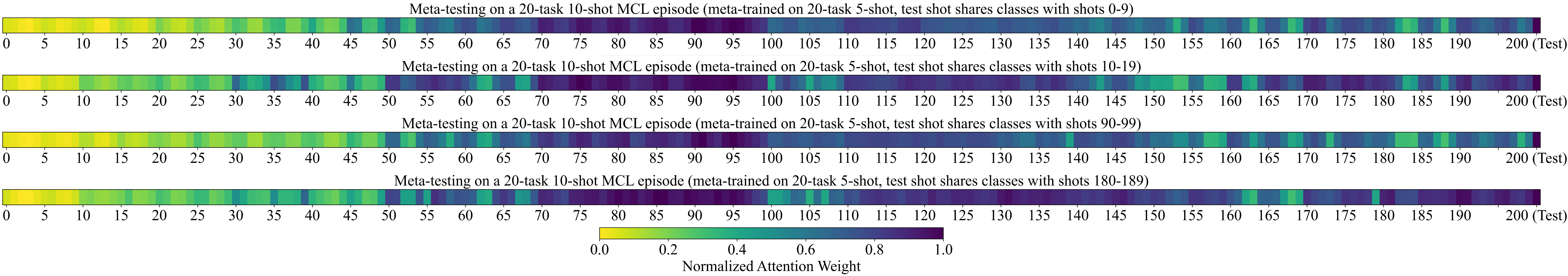}
    \caption{{Transformer}}
    \end{subfigure}
    \caption{\rev{More shots in meta-testing: final-layer associations between test shots (queries) and a single MCL training episode (prompt) for (a) Mamba and (b) Transformer, meta-trained on 20-task 5-shot and meta-tested on 20-task 10-shot episodes. The four visualizations share one training episode spanning the $0^{th}\!\!-\!199^{th}$ shots, and the queries correspond to the $0^{th}$, $1^{st}$, $9^{th}$, and $18^{th}$ tasks, respectively.}}
    \label{fig:more_shot_appendix}
    
    \end{center}
\end{figure}

\begin{figure}[th]
    \begin{center}
    \begin{subfigure}[b]{\textwidth}
            \includegraphics[width=\linewidth]{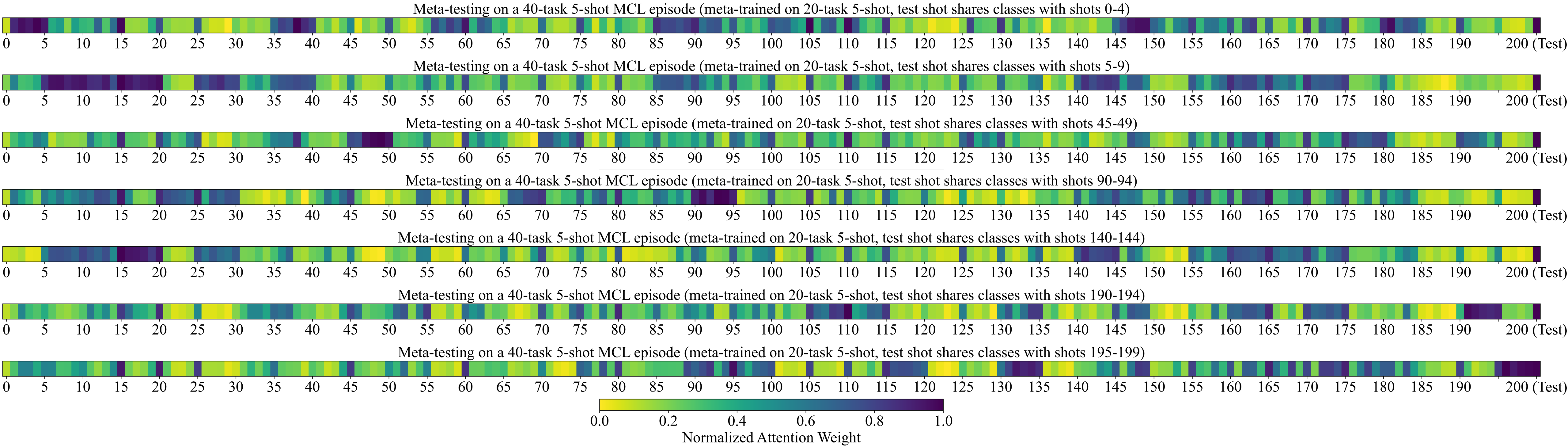}
    \caption{{Mamba}}
    \end{subfigure} \\

    \begin{subfigure}[b]{\textwidth}
            \includegraphics[width=\linewidth]{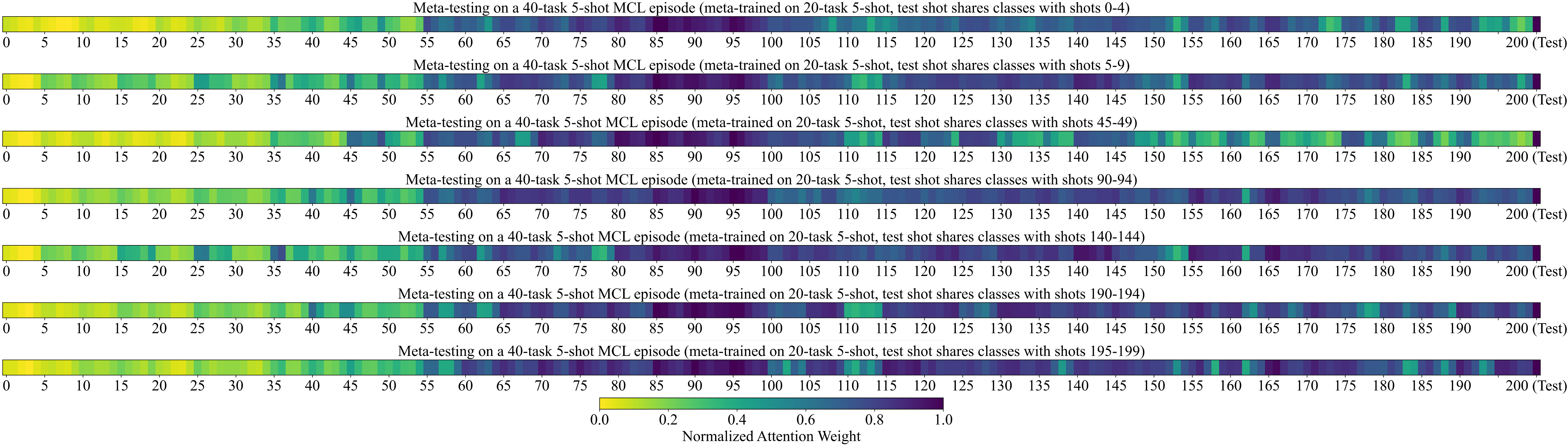}
    \caption{{Transformer}}
    \end{subfigure}
    \caption{\rev{More tasks in meta-testing: final-layer associations between test shots (queries) and a single MCL training episode (prompt) for (a) Mamba and (b) Transformer, meta-trained on 20-task 5-shot and meta-tested on 40-task 5-shot episodes. The seven visualizations share one training episode spanning the $0^{th}\!\!-\!199^{th}$ shots, and the queries correspond to the $0^{th}$, $1^{st}$, $9^{th}$, $18^{th}$, $28^{th}$, $38^{th}$, and $39^{th}$ tasks, respectively.}}
    \label{fig:more_task__appendix}
    
    \end{center}
\end{figure}

\rev{We visualize the weights on 20-task 5-shot (Fig.~\ref{fig:20t5s_appendix}), 20-task 10-shot (Fig.~\ref{fig:more_shot_appendix}), and 40-task 5-shot (Fig.~\ref{fig:more_task__appendix}) meta-testing episodes. On episodes that deviate from the meta-training length, the Transformer tends to either spread its attention uniformly or fixate on specific token positions, whereas Mamba still associates the query with the relevant shots. This suggests that Transformers pick up pattern biases in the sequences (\eg, positional biases unrelated to content), which leads to meta-overfitting in these generalization tests.}

\subsection{Visualization Analysis of Generalization to Noise-Contaminated Episodes}

\begin{figure}[th]
    \begin{center}
    \begin{subfigure}[b]{\textwidth}
            \includegraphics[width=\linewidth]{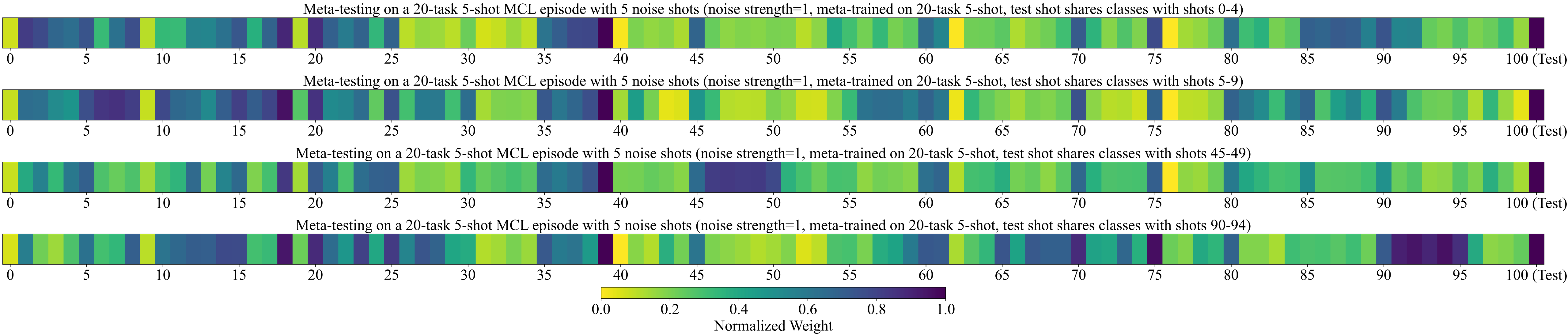}
    \caption{{Mamba}}
    \end{subfigure} \\

    \begin{subfigure}[b]{\textwidth}
            \includegraphics[width=\linewidth]{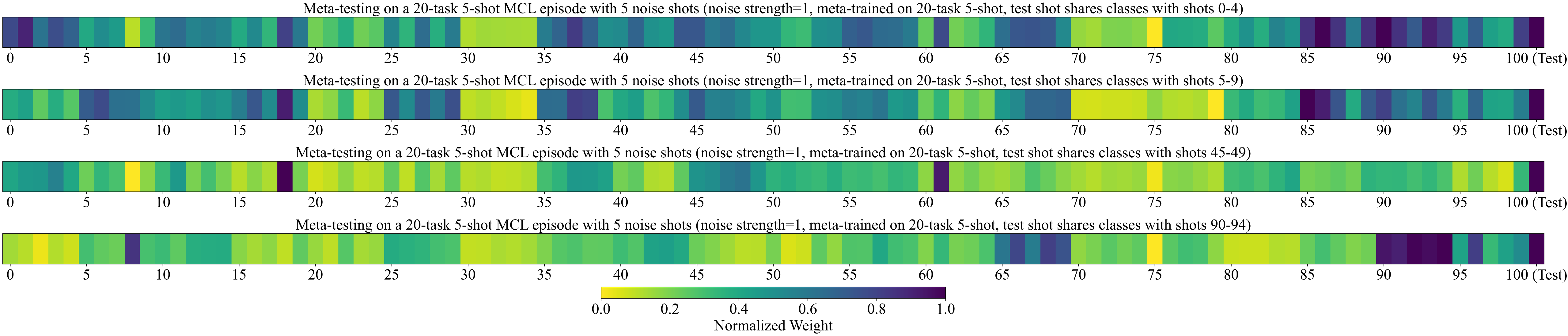}
    \caption{{Transformer}}
    \end{subfigure}
    \caption{\rev{Noise inputs in meta-testing (strength 1): final-layer associations between test shots (queries) and a single MCL training episode (prompt) for (a) Mamba and (b) Transformer, meta-trained on 20-task 5-shot without noise and meta-tested on 20-task 5-shot episodes with noise added to the $8^{th}$, $18^{th}$, $39^{th}$, $61^{st}$, and $75^{th}$ training shots. The four visualizations share one training episode spanning the $0^{th}\!\!-\!99^{th}$ shots, and the queries correspond to the $0^{th}$, $1^{st}$, $9^{th}$, and $18^{th}$ tasks (the $0^{th}\!\!-\!4^{th}$, $5^{th}\!\!-\!9^{th}$, $45^{th}\!\!-\!49^{th}$, and $90^{th}\!\!-\!94^{th}$ training shots), respectively.}}
    \label{fig:noise_1_appendix}
    
    \end{center}
\end{figure}

\begin{figure}[th]
    \begin{center}
    \begin{subfigure}[b]{\textwidth}
            \includegraphics[width=\linewidth]{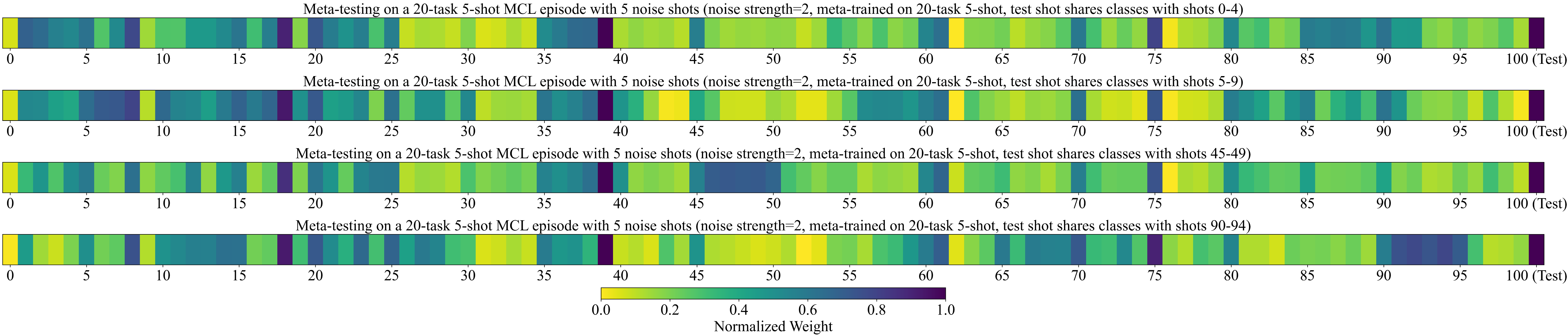}
    \caption{{Mamba}}
    \end{subfigure} \\

    \begin{subfigure}[b]{\textwidth}
            \includegraphics[width=\linewidth]{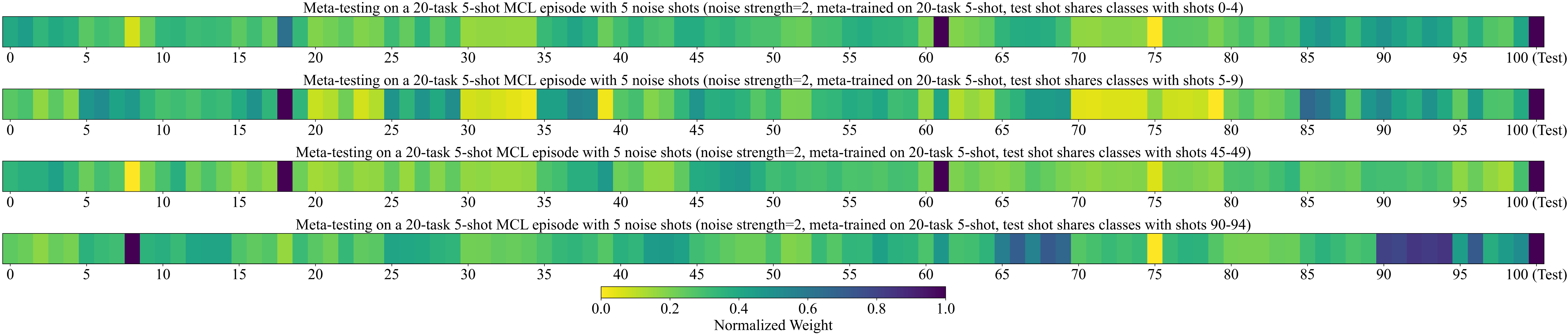}
    \caption{{Transformer}}
    \end{subfigure}
    \caption{\rev{Noise inputs in meta-testing (strength 2): final-layer associations between test shots (queries) and a single MCL training episode (prompt) for (a) Mamba and (b) Transformer, meta-trained on 20-task 5-shot without noise and meta-tested on 20-task 5-shot episodes with noise added to the $8^{th}$, $18^{th}$, $39^{th}$, $61^{st}$, and $75^{th}$ training shots. The four visualizations share one training episode spanning the $0^{th}\!\!-\!99^{th}$ shots, and the queries correspond to the $0^{th}$, $1^{st}$, $9^{th}$, and $18^{th}$ tasks (the $0^{th}\!\!-\!4^{th}$, $5^{th}\!\!-\!9^{th}$, $45^{th}\!\!-\!49^{th}$, and $90^{th}\!\!-\!94^{th}$ training shots), respectively.}}
    \label{fig:noise_2_appendix}
    
    \end{center}
\end{figure}

\begin{figure}[!th]
    \begin{center}
    \begin{subfigure}[b]{\textwidth}
            \includegraphics[width=\linewidth]{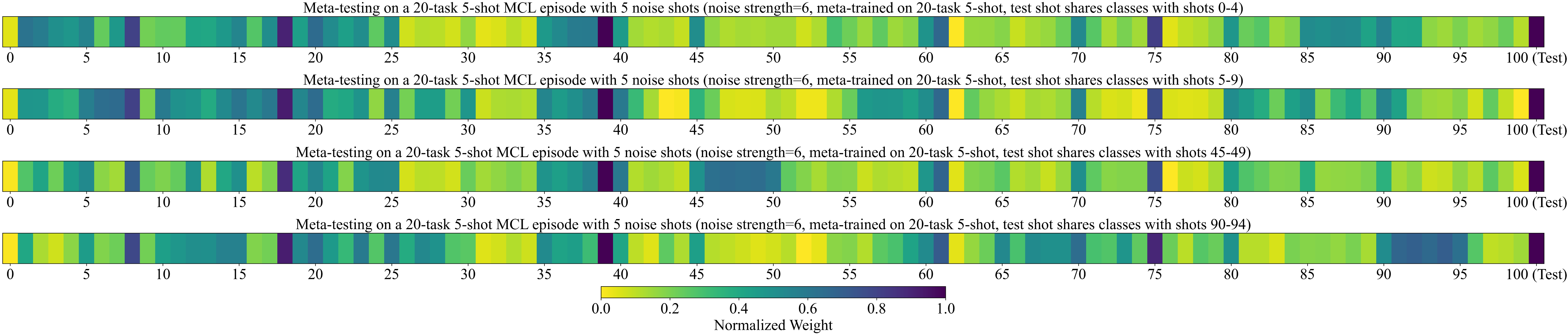}
    \caption{{Mamba}}
    \end{subfigure} \\

    \begin{subfigure}[b]{\textwidth}
            \includegraphics[width=\linewidth]{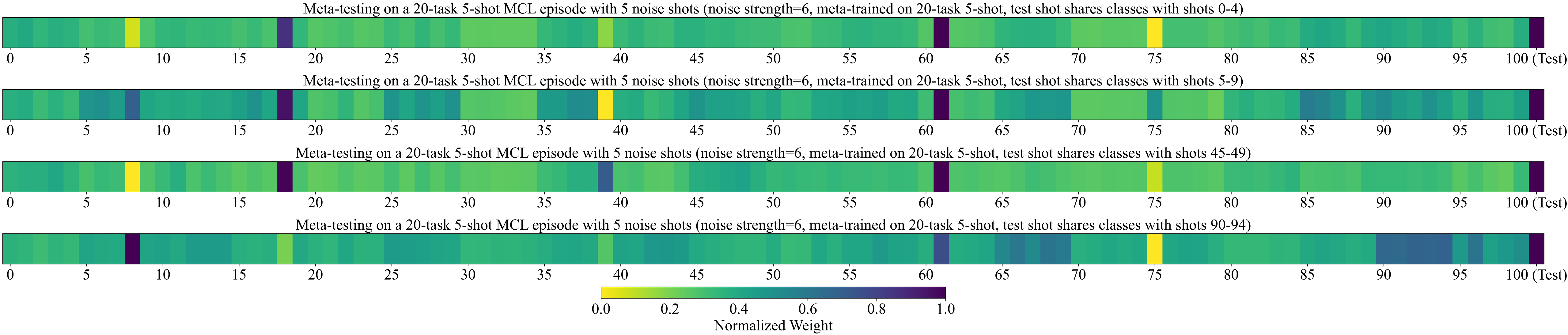}
    \caption{{Transformer}}
    \end{subfigure}
    \caption{\rev{Noise inputs in meta-testing (strength 6): final-layer associations between test shots (queries) and a single MCL training episode (prompt) for (a) Mamba and (b) Transformer, meta-trained on 20-task 5-shot without noise and meta-tested on 20-task 5-shot episodes with noise added to the $8^{th}$, $18^{th}$, $39^{th}$, $61^{st}$, and $75^{th}$ training shots. The four visualizations share one training episode spanning the $0^{th}\!\!-\!99^{th}$ shots, and the queries correspond to the $0^{th}$, $1^{st}$, $9^{th}$, and $18^{th}$ tasks (the $0^{th}\!\!-\!4^{th}$, $5^{th}\!\!-\!9^{th}$, $45^{th}\!\!-\!49^{th}$, and $90^{th}\!\!-\!94^{th}$ training shots), respectively.}}
    \label{fig:noise_6_appendix}
    
    \end{center}
\end{figure}

\rev{In these experiments, the models are meta-trained on noise-free episodes, and noise is added to randomly selected shots of the meta-testing episodes; the task can thus be seen as testing the ability to ignore irrelevant or contaminated outlier samples in the stream. We visualize the final-layer attention/association weights between the test shots and the training shots for both models (meta-trained on the 20-task 5-shot setting), on a 20-task 5-shot episode with five noisy input shots (shot indices 8, 18, 39, 61, and 75) at noise strengths of 1 (Fig.~\ref{fig:noise_1_appendix}), 2 (Fig.~\ref{fig:noise_2_appendix}), and 6 (Fig.~\ref{fig:noise_6_appendix}). The Transformer, meta-trained on clean episodes, tends to produce extreme attention weights (either very high or very low) on the noisy shots, whereas Mamba is much less affected. This suggests that the Transformer's learned attention associates samples through local, independent representations, while Mamba selectively associates relevant information through its recurrently updated state, which accumulates global sequence information.}

\raggedbottom

\clearpage

\end{document}